\documentclass[10pt,conference]{IEEEtran}
\usepackage{cite}
\usepackage{amsmath,amssymb,amsfonts}
\usepackage{algorithmic}
\usepackage{graphicx}
\usepackage{textcomp}
\usepackage[usenames,dvipsnames]{xcolor}
\usepackage[hyphens]{url}
\usepackage{fancyhdr}
\usepackage{hyperref}

\usepackage[normalem]{ulem}
\usepackage{color}
\usepackage{soul}
\usepackage{comment}
\usepackage{booktabs}
\usepackage{enumitem}
\usepackage{listings}
\usepackage{subcaption}
\usepackage{tikz}
\hypersetup{
	unicode=true,
        bookmarksnumbered=true,
	colorlinks=true,
	linkcolor={red!70!black},
	citecolor={red!70!black},
	urlcolor={blue!70!black},
	pdfborder={0 0 0}
}
\usepackage[capitalize,nameinlink]{cleveref}
\usepackage{flushend}
\usepackage{mathtools}
\DeclarePairedDelimiter{\ceil}{\lceil}{\rceil}

\newcommand{\system}{DynamoLLM}
\newcommand{\provider}{\emph{Azure}}

\newcommand*{\rom}[1]{\uppercase\expandafter{\romannumeral #1\relax}}

\usepackage{tcolorbox}
\usepackage{tabularx}
\usepackage{array}
\usepackage{colortbl}
\tcbuselibrary{skins}

\newcolumntype{Y}{>{\raggedleft\arraybackslash}X}

\tcbset{tab1/.style={fonttitle=\bfseries\large,fontupper=\normalsize\sffamily,
colback=yellow!10!white,colframe=red!75!black,colbacktitle=Salmon!40!white,
coltitle=black,center title,freelance,frame code={
\foreach \n in {north east,north west,south east,south west}
{\path [fill=red!75!black] (interior.\n) circle (3mm); };},}}

\tcbset{tab2/.style={enhanced,fonttitle=\bfseries,fontupper=\normalsize\sffamily,
colback=yellow!10!white,colframe=red!50!black,colbacktitle=Salmon!40!white,
coltitle=black,center title}}

\begin{document}

\title{\system{}: Designing LLM Inference Clusters for Performance and Energy Efficiency}

\newcommand{\conferencepubid}{0000--0000/00\$00.00}
\newcommand\conferenceauthors{First Author$\dagger$ and Second Author$\ddagger$}
\newcommand\conferenceaffiliation{First Affiliation$\dagger$, Second Affiliation$\ddagger$}
\newcommand\conferenceemail{Email(s)}


\author{
    \author{Jovan Stojkovic, Chaojie Zhang\textsuperscript{\textdagger},  Íñigo Goiri\textsuperscript{\textdagger}, Josep Torrellas, Esha Choukse\textsuperscript{\textdagger}\\ \textit{University of Illinois at Urbana-Champaign} \qquad \textsuperscript{\textdagger}\textit{Microsoft Azure Research - Systems}\\ }
}

\fancypagestyle{camerareadyfirstpage}{%
  \fancyhead{}
  \renewcommand{\headrulewidth}{0pt}
  \fancyhead[C]{
    \ifdefined\aeopen
    \parbox[][12mm][t]{13.5cm}{\conferenceyear{} IEEE International Symposium}    
    \else
      \ifdefined\aereviewed
      \parbox[][12mm][t]{13.5cm}{\conferenceyear{} IEEE International Symposium}
      \else
      \ifdefined\aereproduced
      \parbox[][12mm][t]{13.5cm}{\conferenceyear{} IEEE International Symposium}
      \else
      \parbox[][0mm][t]{13.5cm}{\conferenceyear{} IEEE International Symposium}
    \fi 
    \fi 
    \fi 
    \ifdefined\aeopen 
      \includegraphics[width=12mm,height=12mm]{ae-badges/open-research-objects.pdf}
    \fi 
    \ifdefined\aereviewed
      \includegraphics[width=12mm,height=12mm]{ae-badges/research-objects-reviewed.pdf}
    \fi 
    \ifdefined\aereproduced
      \includegraphics[width=12mm,height=12mm]{ae-badges/results-reproduced.pdf}
    \fi
  }
  \fancyfoot[C]{}
}
\fancyhead{}
\renewcommand{\headrulewidth}{0pt}

\maketitle

\ifdefined\conferencecameraready 
  \thispagestyle{camerareadyfirstpage}
  \pagestyle{empty}
\else
  \thispagestyle{plain}
  \pagestyle{plain}
\fi

\newcommand{\conferenceheight}{0mm}
\ifdefined\eaopen
\renewcommand{\conferenceheight}{12mm}
\fi


\begin{abstract}

The rapid evolution and widespread adoption of
generative large language models (LLMs) 
have made them a pivotal workload in various applications. 
Today, LLM inference clusters receive a large number of queries with strict Service Level Objectives (SLOs).
To achieve the desired performance, these models execute on power-hungry GPUs causing the inference clusters to consume large amount of energy and, consequently, result in excessive carbon emissions.
Fortunately, we find that there is a great opportunity to exploit the \emph{heterogeneity} in 
inference compute properties and \emph{fluctuations} in
inference workloads, to significantly improve energy-efficiency.
However, such a diverse and dynamic environment creates a large search-space where different system configurations (\emph{e.g.}, number of instances, model parallelism, and GPU frequency) translate into different energy-performance trade-offs.
To address these challenges, we propose {\em \system}, the first energy-management framework for LLM inference environments.
\system{} automatically and dynamically reconfigures the inference cluster to optimize for energy and cost of LLM serving under the service's performance SLOs. 
We show that at a service-level, \system{} 
conserves 53\% energy and 38\% operational carbon emissions, and reduces 61\% cost to the customer, while meeting the latency SLOs.

\end{abstract}

\section{Introduction}
\label{sec:intro}

The exponential growth in the adoption of generative large language models (LLMs) has positioned them at the core of numerous technological advancements and applications.
Today, we see use-cases of LLMs in various domains, such as healthcare~\cite{llmHealth}, 
developer productivity~\cite{copilot},
data analytics~\cite{analytics}, education~\cite{education} and other.
As the popularity of LLMs increases among users, the inference clusters receive millions of queries per day~\cite{statChat} resulting in large infrastructures with 
sophisticated software and expensive hardware systems.

To meet these ever increasing computing demands,
researchers proposed various software~\cite{orca,vllm,flashattention,spotserve,pets} and hardware~\cite{splitwise,alizadeh2024llm,alisa} techniques. Such techniques improve the performance efficiency of LLM inference clusters.
However, one aspect that has been largely overlooked is the energy consumption of these environments~\cite{wordstowatts,towardsGreen}.
The substantial energy requirements of serving LLMs running on power-hungry GPUs have emerged as a significant concern.
As these models become integral to various services, minimizing their energy consumption and, consequently, carbon emissions while maintaining high performance is paramount.

To address this gap, this paper 
starts by characterizing the energy-efficiency properties of LLM inference workloads.
Our characterization underscores
that such environments present a distinct set of challenges, divergent from existing energy management schemes tailored for traditional datacenters applications~\cite{rubik,gemini,pegasus,ecofaas,adrenaline,retail}.
Specifically, we observe that \emph{heterogeneity} in LLM inference compute properties and \emph{fluctuations} in LLM inference workloads
create a dynamic environment with large variations.
Such variations arise from:
(1) requests with varying input/output token lengths,
(2) distinct compute properties of different LLMs,
and (3) different SLOs required by the services using an LLM.

Requests with a large number of input tokens are compute intensive, thus, sensitive to GPU frequency.
Conversely, requests with a few input tokens and many output tokens have low compute, but high memory requirements. Reducing their GPU frequency would save the energy without significantly impacting the performance.
Moreover, 
the number of model parameters also affects the LLM's sensitivity to
the number of GPUs and GPU frequency.
Finally, depending on the service currently using the LLM, the SLO requirements can be strict requiring high-performance configurations,
or loose allowing for lower-performance but more energy-efficient configurations.
Importantly, these characteristics rapidly change due to load fluctuations and dynamic distributions of requests. Such dynamic changes 
cause a system configuration that is energy-efficient at a given point, to quickly become sub-optimal.
This requires a \emph{dynamic approach} to resource management.

To 
pave the way towards energy-efficient and sustainable LLM inference clusters,
this paper introduces {\em \system}, the first energy-management framework for
LLM inference environments. 
\system{} exploits the unique properties of LLM inference workloads to reduce their energy consumption while meeting the performance SLOs.
The system uses energy-performance profiles of models and their workloads to \emph{automatically} and \emph{dynamically} select the energy-efficient configuration.
It leverages multiple 
knobs, including
scaling in/out the number of server instances, model parallelism across GPUs, and GPU frequency scaling.

To handle workload heterogeneity, \system{} maintains differently configured pools of LLM instances that are optimal
for different types of incoming requests.
For instance, compared to a request with many input and output tokens, a request that processes and outputs fewer tokens runs more efficiently on a model parallelized across fewer GPUs running at a lower frequency.
As request distribution varies over time,
\system{} dynamically sizes the pools.
These pools can be merged into fewer pools or divided into multiple pools over time,
providing a balance between right-sizing and fragmentation of resources. 
To efficiently manage the resources,
\system{} 
uses a hierarchy of controllers that reduces computation complexity and eliminates centralized bottlenecks.
The controller at each level operates under the conditions imposed by the upper level, computes its dedicated knob, and forwards further constraints to the controllers at
a lower level. Finally, to enable frequent and smooth transition across different configurations, \system{} includes techniques to minimize or hide the reconfiguration overheads.
As a result, the system maintains high levels of efficiency and service quality under changing workload demands.

An evaluation of  \system{} with a large GPU cluster running production-level traces from a major cloud provider shows that \system{} is very effective:
it
conserves 53\% energy, 38\% operational carbon emissions, and reduces 61\% cost to the customer, while meeting the latency SLOs.

The contributions of this paper are as follows: 
\begin{itemize}[leftmargin=*]
\item An analysis of the opportunities for energy-efficient LLM serving, rooting in the heterogeneity and fluctuations within the inference workloads.

\item Design and implementation of \system{}, a high performance and energy-optimized framework for LLM inference.

\item An evaluation of \system{} on a large-scale platform using production-level traces.
 \end{itemize}

\begin{table*}[t]
\centering
\footnotesize
\begin{tabular}{cccccccccccccc}
\toprule
\multicolumn{2}{l}{Tensor Parallelism} & \multicolumn{4}{c}{TP2} & \multicolumn{4}{c}{TP4} & \multicolumn{4}{c}{TP8}\\
\multicolumn{2}{l}{GPU Frequency (GHz)} &
0.8 & 1.2 & 1.6 & 2.0 &
0.8 & 1.2 & 1.6 & 2.0 &
0.8 & 1.2 & 1.6 & 2.0 \\
Input & Output \\
\midrule
Short & Short & \cellcolor{gray} &	\cellcolor{green!30} \textbf{0.77} &	\cellcolor{orange!50}0.97 &	\cellcolor{orange!90}1.03 &	\cellcolor{orange!30}0.94 &	\cellcolor{yellow!50}0.79 &	\cellcolor{yellow!80}0.91 &	\cellcolor{orange!70}1.01 &	\cellcolor{red!55}1.35 &	\cellcolor{red!25}1.19 &	\cellcolor{red!35}1.29 &	\cellcolor{red!70}1.49\\
Short & Medium & \cellcolor{gray} &	\cellcolor{green!30} \textbf{2.78} &	\cellcolor{orange!50}3.45 &	\cellcolor{orange!70}3.68 & \cellcolor{orange!30}3.39 &	\cellcolor{yellow!50}2.82 &	\cellcolor{yellow!80}3.37 &	\cellcolor{orange!90}3.81 &	\cellcolor{red!55}4.55 &	\cellcolor{red!25}4.15 &	\cellcolor{red!35}4.43 &	\cellcolor{red!70}4.74 \\
Short & Long & \cellcolor{gray} &	\cellcolor{gray}	& \cellcolor{gray}	 & \cellcolor{gray} &	\cellcolor{yellow!50}4.84 &	\cellcolor{green!30}\textbf{4.17} &	\cellcolor{yellow!80}4.97 & \cellcolor{orange!30}5.52 &	\cellcolor{red!35}6.37 &	\cellcolor{orange!70}5.62 & \cellcolor{orange!50}5.59 &	\cellcolor{red!70}6.95\\
Medium & Short & \cellcolor{gray} &	\cellcolor{gray} &	\cellcolor{green!30}\textbf{1.02} &	\cellcolor{orange!70}1.09 &	\cellcolor{gray}	& \cellcolor{orange!50}1.08 &	\cellcolor{orange!30}1.07 &	\cellcolor{orange!90}1.20 &	\cellcolor{red!55}1.51 &	\cellcolor{red!25}1.29 &	\cellcolor{red!35}1.34 &	\cellcolor{red!70}1.73  \\
Medium & Medium & \cellcolor{gray} &	\cellcolor{gray} &	\cellcolor{gray} &	\cellcolor{gray} & \cellcolor{gray} & \cellcolor{orange!30}4.23 &	\cellcolor{green!30}\textbf{3.91} &	\cellcolor{yellow!80}4.08 &	\cellcolor{red!35}5.34 &	\cellcolor{orange!50}4.39 &	\cellcolor{orange!70}4.56 &	\cellcolor{red!70}5.44\\
Medium & Long & \cellcolor{gray} &	\cellcolor{gray} & \cellcolor{gray} & \cellcolor{gray} &	\cellcolor{gray} &	\cellcolor{orange!30}4.99 &	\cellcolor{yellow!80}4.66 &	\cellcolor{green!30}\textbf{4.53} &	\cellcolor{red!35}6.86 &	\cellcolor{orange!50}5.79 &	\cellcolor{orange!70}6.52 &	\cellcolor{red!70}7.12\\
Long & Short & \cellcolor{gray} &	\cellcolor{gray} &	\cellcolor{gray} &	\cellcolor{gray} &	\cellcolor{gray} &	\cellcolor{green!30}\textbf{1.51} &	\cellcolor{yellow!80}1.64 &	\cellcolor{orange!30}1.76 &	\cellcolor{orange!70}2.55 &	\cellcolor{orange!50}2.53 &	\cellcolor{red!35}2.83 &	\cellcolor{red!70}2.94\\
Long & Medium & \cellcolor{gray}	& \cellcolor{gray}	 & \cellcolor{gray}	& \cellcolor{gray} &	\cellcolor{gray} &	\cellcolor{gray} &	\cellcolor{gray} &	\cellcolor{gray} &	\cellcolor{gray}	 & \cellcolor{green!30}\textbf{7.71} &	\cellcolor{orange!30}8.81 &	\cellcolor{red!70}9.17\\
Long & Long & \cellcolor{gray} &	\cellcolor{gray} & \cellcolor{gray} & \cellcolor{gray} & \cellcolor{gray} & \cellcolor{gray} & \cellcolor{gray} & \cellcolor{gray} & \cellcolor{gray} & \cellcolor{orange!30}12.99 &	\cellcolor{green!30}\textbf{11.89} &	\cellcolor{red!70}13.21\\
\bottomrule
\end{tabular}
\caption{Energy consumption in Watt$\times$hours (Wh) for Llama2-70B varying request lengths, frequency, and model parallelism with medium system load (2K tokens per second). 
Configurations that violate the SLO are shown as empty gray boxes, while the acceptable configurations are colored as a heat map according to their energy consumption, per row.}
\label{tab:energy_request_types}
\end{table*}

\section{Background}
\label{sec:background}

\vspace{1pt}
\noindent\textbf{Computational phases of LLMs}
Generative LLMs~\cite{llama2, llama3, mixtral2, gemini, radford2019gpt} are auto-regressive:
they process the whole input in parallel, and serially generate the output tokens.
This property leads to two computationally distinct phases~\cite{polca,splitwise}.
First is the prefill phase, where the input tokens are computed in parallel.
This is a compute-intensive phase and scales with the number of input tokens. 
Second is the decode phase, where each output token is generated serially, based on all the tokens seen so far.
This is a memory-intensive phase, and scales with the number of output tokens.

\vspace{1pt}
\noindent\textbf{Performance metrics for LLMs}
To evaluate the performance, we use:
time to first token (TTFT), time between tokens (TBT), and throughput ~\cite{splitwise,guideMetrics}.
TTFT is the latency of generating the first output token;
while TBT is the latency to generate each new output token.
To quantify the energy efficiency, 
we measure the energy consumption 
in Watt-hours (Wh) while meeting certain latency SLOs. 
The SLOs 
vary depending on their use cases for different tasks. For latency-sensitive tasks, both TTFT and TBT are important metrics with strict SLOs.
We define SLOs for TTFT and TBT based on maximum achievable performance, described in \Cref{tab:trace_conf_in_out_tokens}.

\vspace{1pt}
\noindent\textbf{LLM parallelism}
A single model can be divided across GPUs
to improve performance and allow larger memory footprints.
LLM inference typically uses pipeline and tensor parallelism.
Pipeline parallelism (PP) partitions the LLM layers among GPUs, while keeping all the operators/tensors of a layer on the GPU.
GPUs then communicate only in between two consecutive stages.
Tensor parallelism (TP) allocates a slice of each layer to each GPU.
This requires aggregation across all the GPU for each layer, in turn needing high bandwidth communication.
TP performs better for GPUs within the same server, connected with high bandwidth interconnects (\emph{e.g.}, NVLink~\cite{nvlink}), while PP is preferred across servers.
Since most open source models~\cite{llama2, llama3, mixtral2} fit on 8 GPUs in a single server, we consider only TP in the rest of the paper; the ideas can easily extend to PP.
We denote tensor parallelism across 2, 4 and 8 GPUs as TP2, TP4 and TP8, respectively.

\vspace{1pt}
\noindent\textbf{Power and energy in datacenters}
A rich body of work explored power/energy efficiency in traditional datacenters~\cite{pegasus,thunderbolt,smartoclock,retail}. 
However, the rapid growth of LLMs 
has posed new challenges 
that have not yet been extensively studied. 
LLM inference workloads comprise a swiftly increasing percentage of datacenter load~\cite{polca}.
This, coupled with the power-dense hardware like DGX A100s and H100s being deployed to serve these workloads makes them power, energy, and carbon-intensive~\cite{polca,wordstowatts,llmcarbon}.
To effectively address this challenge,
it is important to have a comprehensive framework for managing energy in these systems.

\begin{table*}[t]
\centering
\footnotesize
\begin{tabular}{ccccccccccccc}
\toprule
Tensor Parallelism & \multicolumn{4}{c}{TP2} & \multicolumn{4}{c}{TP4} & \multicolumn{4}{c}{TP8}\\
GPU Frequency (GHz) &
0.8 & 1.2 & 1.6 & 2.0 &
0.8 & 1.2 & 1.6 & 2.0 &
0.8 & 1.2 & 1.6 & 2.0\\
\midrule
Low Load & \cellcolor{gray} & \cellcolor{gray}	&	\cellcolor{yellow!50}3.41 &	\cellcolor{orange!70}3.75 &	\cellcolor{yellow!80}3.44 &	\cellcolor{green!30}\textbf{2.93} &	\cellcolor{orange!30}3.71 &	\cellcolor{orange!50}3.73 &	\cellcolor{red!35}4.49 &	\cellcolor{red!25}3.76 & \cellcolor{red!55}4.52 & \cellcolor{red!70}4.64
\\
Medium Load & \cellcolor{gray} &	\cellcolor{gray} &	\cellcolor{gray} &	\cellcolor{gray} & \cellcolor{gray} & \cellcolor{orange!30}4.23 &	\cellcolor{green!30}\textbf{3.91} &	\cellcolor{yellow!80}4.08 &	\cellcolor{red!35}5.34 &	\cellcolor{orange!50}4.39 &	\cellcolor{orange!70}4.56 &	\cellcolor{red!70}5.44\\
High Load & \cellcolor{gray} &	\cellcolor{gray} &	\cellcolor{gray} &	\cellcolor{gray} &	\cellcolor{gray} &	\cellcolor{gray} & \cellcolor{orange!30}4.22 & 
\cellcolor{green!30}\textbf{4.13} & \cellcolor{red!35}5.86 &	\cellcolor{orange!50}5.24 &	\cellcolor{orange!70}5.42 &	\cellcolor{red!70}6.62\\
\bottomrule
\end{tabular}
\caption{Energy consumption in Wh for LLama2-70B medium-sized input and output (MM)  requests varying frequency and model parallelism under different system loads: low (650 TPS), medium (2K TPS) and high (4K TPS).
}
\label{tab:energy_loads}
\end{table*}

\begin{table*}[t]
\footnotesize
\centering
\begin{tabular}{ccccccccccccc}
\toprule
Tensor Parallelism & \multicolumn{4}{c}{TP2} & \multicolumn{4}{c}{TP4} & \multicolumn{4}{c}{TP8}\\
GPU Frequency (GHz) &
0.8 & 1.2 & 1.6 & 2.0 &
0.8 & 1.2 & 1.6 & 2.0 &
0.8 & 1.2 & 1.6 & 2.0\\
\midrule
Llama2-13B~\cite{llama1} &  
\cellcolor{yellow!50}1.05 &	
\cellcolor{green!30}\textbf{0.99} &	\cellcolor{yellow!70}1.14 &	\cellcolor{yellow!90}1.24 &	\cellcolor{orange!40}1.52 &	\cellcolor{orange!30}1.27 &	\cellcolor{orange!60}1.58 &	\cellcolor{orange!70}1.65 &	\cellcolor{red!35}2.61 &	\cellcolor{red!25}2.35 &	\cellcolor{red!50}2.74 & \cellcolor{red!70}3.45\\
Mixtral-8x7B~\cite{mixtral1} & \cellcolor{yellow!50}1.03 &	
\cellcolor{green!30}\textbf{0.98} &	\cellcolor{yellow!70}1.21 &	\cellcolor{yellow!90}1.32 &	\cellcolor{orange!30}1.39 &	\cellcolor{orange!40}1.51 &	\cellcolor{orange!60}2.09 &	\cellcolor{orange!70}2.31 &	\cellcolor{red!25}2.57 &	\cellcolor{red!35}3.06 &	\cellcolor{red!50}3.71 & \cellcolor{red!70}4.66\\
Llama2-70B~\cite{llama} & \cellcolor{gray} &	\cellcolor{gray} &	\cellcolor{gray} &	\cellcolor{gray} & \cellcolor{gray} & \cellcolor{orange!30}4.23 &	\cellcolor{green!30}\textbf{3.91} &	\cellcolor{yellow!80}4.08 &	\cellcolor{red!35}5.34 &	\cellcolor{orange!50}4.39 &	\cellcolor{orange!70}4.56 &	\cellcolor{red!70}5.44\\ Llama3-70B~\cite{llama3} & \cellcolor{gray} &	\cellcolor{gray} &	\cellcolor{gray} &	\cellcolor{gray} & \cellcolor{gray} & \cellcolor{yellow!80}4.32 &	\cellcolor{green!30}\textbf{4.28} &	\cellcolor{orange!30}4.57 &	\cellcolor{red!35}6.11 &	\cellcolor{orange!50}5.18 &	\cellcolor{orange!70}5.42 &	\cellcolor{red!70}6.45 \\
Mixtral-8x22B~\cite{mixtral2} & \cellcolor{gray} &	\cellcolor{gray} &	\cellcolor{gray} &	\cellcolor{gray} &	\cellcolor{gray}	& \cellcolor{gray} &	\cellcolor{gray} &	\cellcolor{gray} &
\cellcolor{orange!70}3.83 &	\cellcolor{green!30}\textbf{3.23} &	\cellcolor{yellow!80}3.65 &	\cellcolor{red!70}4.03
\\
Falcon-180B~\cite{falcon} & \cellcolor{gray} &	\cellcolor{gray} &	\cellcolor{gray} &	\cellcolor{gray} &	\cellcolor{gray}	& \cellcolor{gray} &	\cellcolor{gray} &	\cellcolor{gray} &	\cellcolor{orange!70}9.56 &	\cellcolor{green!30}\textbf{7.94} &	\cellcolor{yellow!80}8.57 &	\cellcolor{red!70}10.34 \\
\bottomrule
\end{tabular}
\caption{Energy consumption in Wh for medium-sized (MM) requests of different LLM  architectures varying the frequency and model parallelism with medium system load (2K TPS).
}
\label{tab:energy_models}
\end{table*}

\section{Opportunities for Energy Efficiency}
\label{sec:opportunity}

To understand the energy-efficiency properties of LLM inference environments, we characterize open-source models~\cite{llama,falcon,mixtral1,mixtral2}
on an NVIDIA DGX H100 server~\cite{h100} using vLLM~\cite{vllm} inference engine.
We analyze the energy properties of LLMs by varying the request lengths, request load, model, and service SLO.
Additionally, we analyze how the profiled variables change over time in a real-production environment using the invocation traces of two LLM services from \provider{}:
\emph{Coding} and \emph{Conversation}.
The traces include a subset of invocations received by the profiled services
during one week, and  contain the timestamp of the invocation, along with the number of input  and 
output tokens.
These traces are a super-set of our open-source traces for the same services~\cite{splitwise}.

\subsection{Heterogeneous Energy-Performance Profiles}

\vspace{1pt}
\noindent\textbf{Request lengths}
The prefill and decode phases in an LLM inference exhibit distinct execution behaviors (\Cref{sec:background}), suggesting that requests of \emph{different input and output lengths} possess different compute and energy characteristics.
We categorize the requests based on the number of input/output tokens into 9 buckets:
SS (short input, short output), SM (short input, medium output), SL (short input, long output), MS, MM, ML, LS, LM, and LL.
\Cref{tab:trace_conf_in_out_tokens} shows the thresholds and corresponding TTFT/TBT SLOs.
We set the thresholds for request lengths using the $33^{rd}$, $66^{th}$ and $100^{th}$ percentiles of the input/output lengths from a trace for a \emph{Conversation} service from \provider{}.
We set the SLOs to $5\times$ the latency of a single request running isolated on a system~\cite{alpaserve}.

\begin{table}[t]
\vspace{0mm}
    \centering
    \footnotesize
    \begin{tabular}{ccrrrr}
        \toprule
               &   & Input & Output & TTFT SLO & TBT SLO\\
        \midrule
        Short  & S &  $<$256 & $<$100    &  250 ms & 100 ms\\
        Medium & M & $<$1024 & $<$350    &  400 ms & 100 ms\\
        Long   & L & $\le$8192 & $\ge$350 & 2000 ms & 100 ms\\
        \bottomrule
    \end{tabular}
    \caption{Thresholds for classifying the requests based on input/output lengths and corresponding TTFT/TBT SLOs.
    }
    \label{tab:trace_conf_in_out_tokens}
 \vspace{-2mm}
\end{table}

We use these categories to characterize the energy consumption of different request types running the Llama2-70B~\cite{llama} model with a medium system load of 2000 tokens per second (TPS) under various GPU frequencies and model parallelisms.
\autoref{tab:energy_request_types} shows our results in the form of a heat map.
Since shorter requests are not computationally intensive, they meet their SLOs with any tensor parallelism, and generally at lower frequencies compared to the rest.
As an example, the least-energy configuration for SS requests is TP2 at 1.2 GHz.
Conversely, LL requests can only run with TP8 without violating the SLO.
With TP8, the least-energy configuration for LL requests is 1.6 GHz.
Note that the lowest power configuration that meets SLOs (TP8 at 1.2 GHz), is not the energy-optimal one due to the increased execution time.
Running all the requests together would require the system to run 
with the most constrained SLO configuration, in this case, as per the LL configuration.
This would make the system energy inefficient.

To exploit this heterogeneity for energy-efficiency, 
the system would need to separate requests based on their input/output lengths, and process different request types with different server configurations.
However, on request arrival, the input length is known, but, due to
the auto-regressive LLM nature, the output length is unknown.
Thus, the system needs to predict the output length.
\system{} will rely on prior work that efficiently performs such operation with relatively high precision~\cite{aiops2024qiu,seqPred,s3}, and will have a mechanism to mitigate the impact of occasional mis-predictions.

\vspace{1pt}
\noindent\textbf{Request loads}
In addition to the request length, the incoming load of the LLM inference server drives the compute requirements.
During  periods of low load, the system has a larger SLO slack to exploit and can run the 
requests at low-frequency configurations to save energy.
Conversely, during  periods of high load, the system does not have enough SLO slack, and needs to run at high-frequency configurations.

\autoref{tab:energy_loads} shows the energy consumed when running Llama2-70B medium-sized 
input and output  (MM) requests while varying the number of processed prompt tokens per second (TPS).
The system can run low load with any TP at almost any frequency.
Among all feasible configurations, the lowest-energy  configuration is TP4 with 1.2 GHz.
TP8 requires more GPUs to operate in parallel and, thus, consumes more energy.
TP2 uses fewer GPUs but increases the execution time and forces individual GPUs to operate at high 
frequency to meet SLOs, leading to high energy.
Conversely, under high load, the system cannot operate on TP2 and requires TP4 or TP8.
The lowest energy  configuration is TP4 with 2 GHz. Overall,
to minimize the energy consumption while operating under performance constraints, we need to consider the incoming load to set the correct  parallelism and GPU frequency.

\vspace{1pt}
\noindent\textbf{Requested model}
The diversity of the compute properties of an LLM directly translates into its energy profile.
\autoref{tab:energy_models} shows the energy consumption of different LLMs when running medium-sized requests at medium system load.
Smaller models, such as Llama2-13B and Mixtral-8x7B, can run with any TP (even with a single GPU); their lowest-energy  configuration is TP2 at 1.2 GHz.
Mixtral-8x22B and Falcon-180B are much larger and can only run with TP8.
Their lowest-energy configuration is TP8 at 1.2 GHz.

Compute-bound models with large number of parameters are more sensitive to the GPU frequency and model parallelism. Hence, they often need to operate at high-frequency and high-energy modes.
Sparse models with relatively smaller numbers of parameters tolerate lower frequencies and lower model parallelism. Hence, they  meet the performance requirements even with lower-performance modes.

\vspace{1pt}
\noindent\textbf{Service SLO}
Different services often use the same model with different SLO requirements~\cite{infaas}.
As indicated before, we assume an SLO such that the P99 tail latency is within 5$\times$ of the execution time of a request on an unloaded system~\cite{alpaserve}.
However, some services have more relaxed SLOs, at 10$\times$ or even 20$\times$ of a single request execution~\cite{amdahltail,usteal}.
For different SLO requirements, the system may need different energy-optimal configurations.
For example, Table~\ref{tab:energy_request_types} shows that, with strict SLO (5$\times$), short-input long-output sized LLama2-70B requests at medium load have the optimal configuration at TP4 and 1.2GHz.
However, if with loose SLO (10$\times$), the requests may even operate with TP2 at 1.6GHz.

\vspace{1pt}\noindent\textbf{\underline{Insight \#1}}
LLM  workloads are highly heterogeneous in their energy-performance profiles.
To achieve the optimal energy under performance SLOs, different requests (sizes, models and SLOs) need to be processed separately and differently.

\subsection{Dynamic LLM Inference Workloads}

\begin{figure}[t]
\centering
\includegraphics[width=\columnwidth]{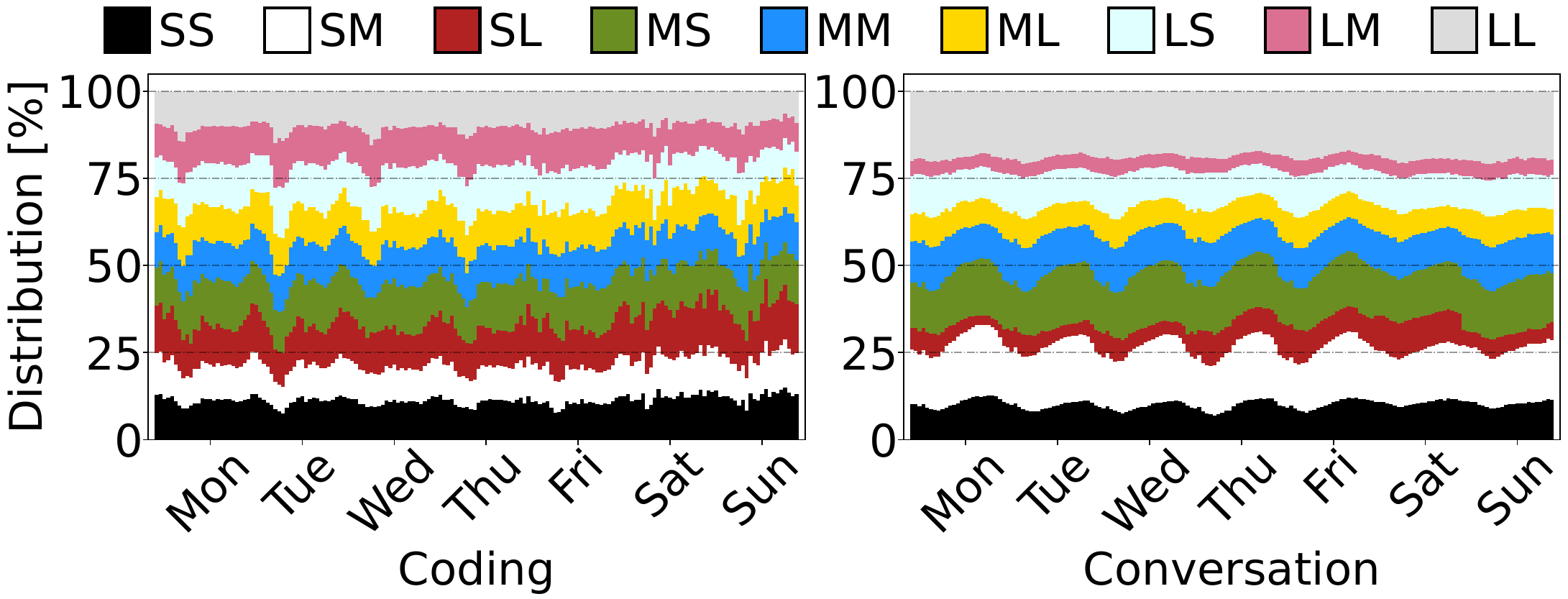}
\caption{Distribution of requests based on input and output lengths categorized into three groups: short, medium, and long.}
\label{fig:char_distribution}
\vspace{-2mm}
\end{figure}

\vspace{1pt}
\noindent\textbf{Changing request-length distribution}
We measure the distribution of request types for \emph{Coding}  and \emph{Conversation} services.
\autoref{fig:char_distribution} shows the distribution of requests for each workload over a week.
The distribution 
differs across services.
 \emph{Conversation} has typically longer  outputs and shorter inputs, while  \emph{Coding} shows the opposite trend.
However,  both services have a significant fraction of each request type, and importantly, the popularity of 
request types changes over time.

As observed earlier, 
different request types require different energy-optimal configurations.
Thus,
the system needs to split its resources into per request-type pools, configure pools individually,
and dynamically adapt the pools' configurations based on the current request distribution.
However, if the system classifies the requests into too few classes, it will not be able to fine-tune the system for best energy.
On the other hand, too many classes may lead to fragmentation and negatively impact  energy efficiency.
Thus, the system has to find the right number of resource pools.
In \system{}, we will use historical data to set the number of pools such that requests with distinct SLO requirements (TTFT or TBT bound) and compute properties (compute or memory bound) have separate pools.
Moreover, as the load of a given request type reduces, \system{} will avoid fragmentation by merging the pool with the next available pool that serves longer requests.

\begin{figure}[t]
\centering
\includegraphics[width=\columnwidth]{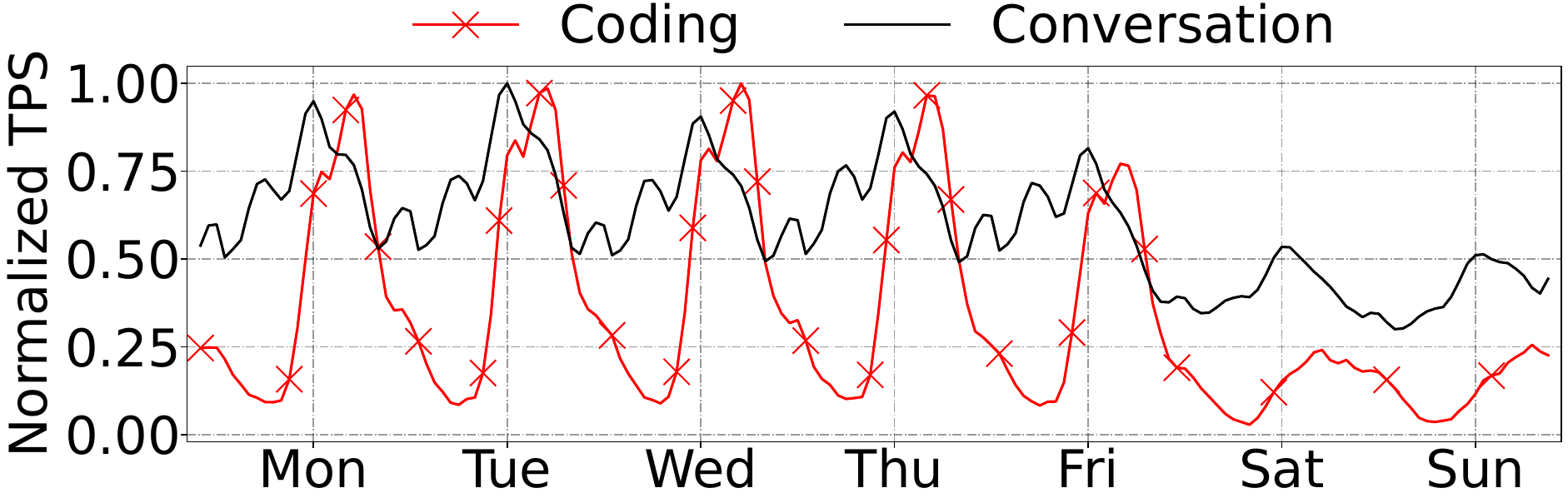}
\caption{Load over a week for  \emph{Coding}  and  \emph{Conversation}  LLM inference workloads.}
\label{fig:char_arrival}
\vspace{-2mm}
\end{figure}

\vspace{1pt}
\noindent\textbf{Request load fluctuations}
 LLM inference workloads, as user-facing applications, exhibit a typical diurnal pattern with peaks during working hours and valleys at night and weekends.
\autoref{fig:char_arrival} shows the load in tokens per second of the two workloads over a week.
The load is normalized to the peak of the individual workloads.
The \emph{Coding} trace shows a clear diurnal pattern, with peaks every day, lower load at night, and much lower load during  weekends.
 \emph{Conversation}  shows a less extreme, but still significant, diurnal pattern.

The peak load of \emph{Conversation} is 1.7$\times$ and 3.3$\times$ higher than its average and valley loads, respectively.
The peak load of \emph{Coding} is 2.8$\times$ and 34.6$\times$ higher than its average and valley loads, respectively.
This large slack indicates that the LLM inference servers can frequently operate in a less performant but energy-optimized configuration  without violating the SLO.
Once the load starts building up, the server needs to switch to a more performant mode of operation.
       
\vspace{1pt}
\noindent\textbf{LLM service SLO and model diversity}
Finally, different services may time-share the same LLM model instance~\cite{oneSize}.
They may have different SLOs, requiring the 
configuration to be adapted based on the current service-user.
On the other hand, the same service may concurrently use multiple different models~\cite{hybridLLM}. This requires different execution plans for the optimal energy consumption of the individual queries.
Thus, it is not trivial for  service providers to operate in an energy-optimal setting while meeting the performance SLOs.

\vspace{1pt}
\noindent\textbf{\underline{Insight \#2}}
LLM workloads are highly dynamic and, thus, an energy-optimal configuration  can quickly become sub-optimal.
However, the complexity of a large search space requires an automatic and user-transparent configuration selection.

\subsection{Reconfiguration Overheads}
To capture the fast changes
in LLM inference workloads, we need to quickly transition between configurations.
However, there are overheads to change (1) number of inference server instances, (2) model parallelism, and (3) GPU frequency.

\vspace{1pt}
\noindent\textbf{Changing instance number}
To adjust to fluctuating load, it is cost-beneficial to dynamically adjust the number of LLM instances to serve the requests (\emph{i.e.}, scale in and out).
However, the overheads of adding a new inference server are too large to be tolerable on the critical path of inference loads.
\Cref{tab:scale} shows the breakdown of the overheads to:
(1) instantiate a new GPU VM in the cloud (such as H100 VM~\cite{azurevm}),
(2) initialize the distributed multi-GPU environment (\emph{e.g.}, Ray, MPI),
(3) download the model weights,
(4) setup the inference engine, and
(5) install the weights and key-value cache on the GPUs.
In total, these overheads can take even a few minutes.
Hence, the conventional LLM inference environments typically provision the static number of instances to handle their peak load resulting in heavy underutilization.
In \system{}, we will propose techniques to efficiently scale the number of instances (with the current load) while minimizing most of the scale-out overheads.

\begin{table}[t]
\centering
\footnotesize
    \begin{tabular}{lc}
    \toprule
    Overhead source  & Time \\
    \midrule
    Create a new H100 VM~\cite{azurevm} & $\sim$1-2 min\\
    Initialize distributed multi-GPU environment & $\sim$2 min\\
    Download model weights (Llama2-70B~\cite{llama2}) & $\sim$3 min\\
    Set up the engine configuration & $\sim$18 sec\\
    Install weights and KV cache on GPUs & $\sim$15 sec \\
    \midrule
    Total & $\sim$6-8 min\\
    \bottomrule
    \end{tabular}
    \caption{Measured overheads of creating a new 8$\times$H100 instance of an LLM inference server VM.
    }
    \label{tab:scale}
\vspace{-2mm}
\end{table}

\vspace{1pt}
\noindent\textbf{Changing model parallelism}
To modify the model parallelism of an LLM inference server, 
we need to perform two operations.
First, we need to re-shard the model weights and transfer them to the memory of the right GPUs.
Second, the inference engine needs to synchronize the involved GPUs.
Current systems stop the engine, unload the weights from GPUs, load the weights from the host to the new set of GPUs, and re-start the engine 
from the scratch.
This adds intolerable overheads (around 1-2 minutes) if performed on the critical path.
In \system{}, we will show how to minimize the re-sharding overheads by smartly mapping the logical to physical GPUs, exploiting inter-GPU direct NVLink connections and moving the weights between GPUs in the background.

\begin{figure}[t]
\centering
\includegraphics[width=\columnwidth]{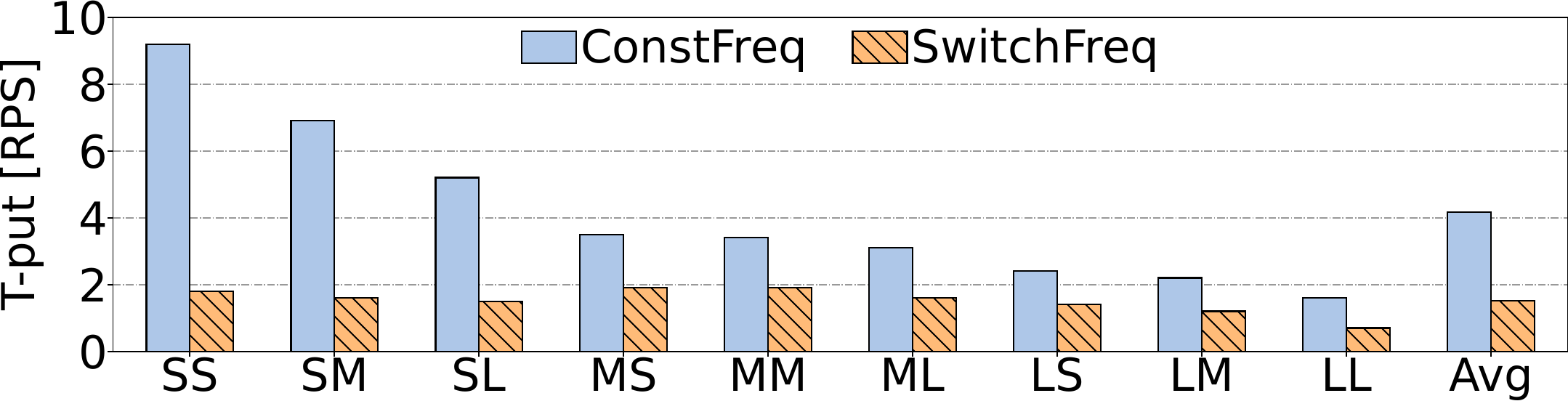}
\caption{Throughput for different request types with constant frequency (1980MHz) and with re-setting the frequency (to 1980MHz) on every iteration in the background.}
\label{fig:char_freqswitch}
\vspace{-5mm}
\end{figure}

\vspace{1pt}
\noindent\textbf{Changing GPU frequency}
Setting the GPU frequency (\emph{e.g.}, via \texttt{nvidia-smi}~\cite{nvidiasmi}) incurs non-negligible overheads.
It involves invoking the OS, communicating with the GPU driver via system calls, and performing hardware interactions via firmware.
On average, setting the GPU frequency takes around 50-80ms.
In comparison, one decode iteration of the LLM inference process takes 20-30ms.
Consequently, the time spent adjusting the GPU frequency can significantly impact the overall performance, potentially doubling the latency of individual inference steps.
\Cref{fig:char_freqswitch} shows the throughput for different request types when constantly running at the highest frequency (1980 MHz) and when re-setting the frequency (to 1980 MHz) in the background on every LLM inference iteration. 
Due to the software overheads, the throughput of LLM inference system significantly drops.
Therefore, optimizing or minimizing frequency changes during LLM inference is crucial for maintaining efficient and responsive performance.

\vspace{1pt}
\noindent\textbf{\underline{Insight \#3}} 
Transitioning between LLM server configurations incurs significant overheads.
For energy-efficiency, such overheads need to be minimized and considered when computing the energy/performance trade-offs. 

\begin{figure*}[t]
    \centering
    \includegraphics[width=2\columnwidth]{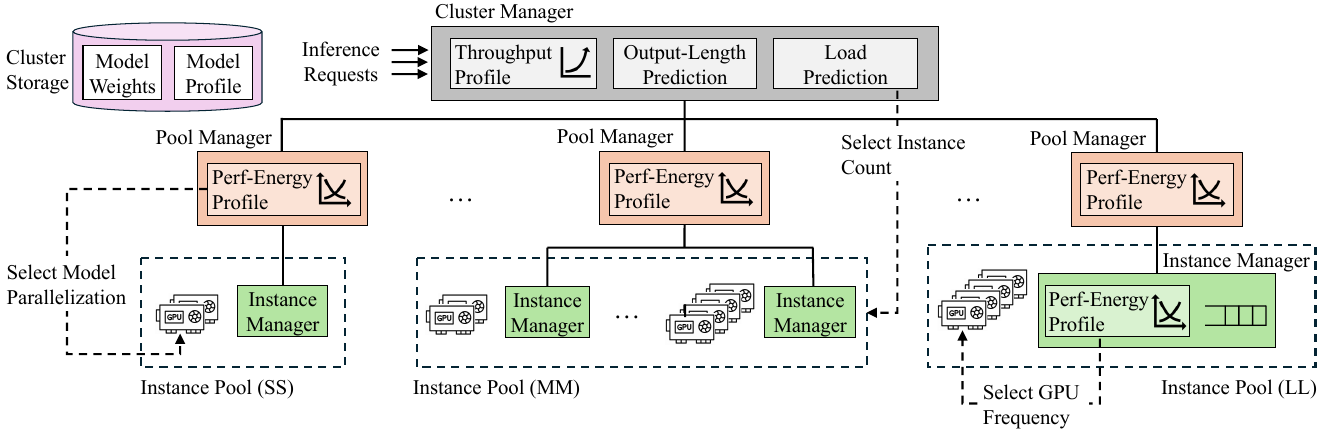}
    \caption{\system{} architecture: a hierarchy of controllers with cluster resources split into per request-type pools.}
    \label{fig:architecture}
    \vspace{-4mm}
\end{figure*}

\section{\system{}: An Energy Management Framework for LLM Inference Clusters}
\label{sec:design}

We use the insights to design \system{}, the first energy management framework for LLM inference environments. 
\system{} seamlessly integrates with existing inference platforms, enabling LLM workloads to operate energy-efficiently and cost-effectively while meeting their performance SLOs. 
\system{} has four key principles.
First, it is \emph{energy-optimized and SLO-aware}, leveraging model profiles to automatically select the most energy-efficient configuration for specific LLMs and inference workloads within their SLO requirements. 
Second, \system{} fine-tunes configurations for heterogeneous LLM workloads by dividing cluster resources into \emph{instance pools} tailored to specific request types. 
Third, \system{} accommodates fluctuating LLM inference loads by \emph{dynamically reconfiguring} the chosen organization. 
Finally, to ensure frequent and smooth reconfiguration, \system{} \emph{minimizes reconfiguration overheads}.

\vspace{1pt}\noindent\textbf{Architecture}
\autoref{fig:architecture} shows the \system{} architecture.
The system is organized hierarchically at the cluster, pool, and instance levels. At each level, the controllers
tune their assigned configuration knob,  and communicate their decisions with the controllers from the upper and lower levels.
The controllers use energy-performance models generated in the profiling phase to determine the number of instances, model parallelization, and GPU frequency for an energy-optimized operation given the current system state.
(1) \emph{Cluster Manager} receives inference requests, predicts their type, and forwards them to the appropriate instance pool.
Additionally, it periodically re-evaluates how many pools and how many model instances per pool are needed based on the system load.
(2) \emph{Pool Manager}
schedules the requests to model instances in a manner that minimizes per-pool energy consumption. It also periodically checks if its instances need to be re-sharded into a more energy-efficient parallelization.
(3) \emph{Instance Manager} schedules the requests to the inference engine and periodically checks if the instance's GPU frequency needs to be adjusted.

\subsection{Configuring Instances for Energy-Efficiency}
\label{sec:energyconfig}

\vspace{1pt}\noindent\textbf{Generating LLM profiles}
When deploying their service to \system{}, users specify the LLM used by the service and the expected performance SLOs. 
Then, the system characterizes the model and generates its energy-performance profile.
\system{} profiles the model by running loads of different request lengths at different model parallelisms (TP2, TP4 and TP8) and GPU frequencies (800--1980MHz, with a step of 200MHz).
The system profiles a few load levels, up to the maximum throughput,
and then extrapolates the behavior for the loads in between the measured ones.
The profiling result is a function that takes the load, request length, model parallelism and GPU frequency as inputs and outputs the expected energy consumption and TTFT/TBT latencies.

As many services may use the same model, \system{} can reuse the profiles across services, minimizing the profiling overheads.
Such profiles are stored in a global \system{} repository, and then cached in a cluster-local storage when a given service is deployed in the cluster.

\vspace{1pt}\noindent\textbf{Selecting the energy-optimized configuration}
Given the current load and available resources, \system{} uses the generated profiles to minimize energy consumption while staying within performance constraints.
The system formulates this task as an optimization problem for the mixed integer linear programming (MILP) solver.
The solver needs to output how many instances of each tensor parallelism ($N^{TP_2}$, $N^{TP_4}$ and $N^{TP_8}$) are needed, at which frequency they should run ($f^{TP_2}$, $f^{TP_4}$ and $f^{TP_8}$),  and which load should be assigned to each instance ($L^{TP_2}$, $L^{TP_4}$ and $L^{TP_8}$).
We assume that all instances of a given parallelism run at the same frequency and receive fair-share amount of work.

The optimization target of the solver is to minimize the total energy consumption ($E$),
while the constraints are:
\emph{1)} the total number of GPUs used by all instance types does not exceed the assigned number of GPUs ($N$);
\emph{2)} the load assigned to individual instances sums up to the total expected load ($L$);
and \emph{3)} the expected performance of all instances with the assigned load is within the requirements ($SLO$).
Functions $Energy^{TP_i,f_i}(L^{TP_i})$ and $Performance^{TP_i, f_i}(L^{TP_i})$ output the expected energy and performance, respectively, when running the load $L^{TP_i}$ with $TP_i$ parallelism at $f_i$ GPU frequency.
Then, the optimization task can be formulated as:
\begin{equation}
\footnotesize
\begin{aligned}
    \min \quad & \left( \sum_{i} (N^{TP_i} \times Energy^{TP_i,f_i}(L^{TP_i})) \right) \quad \forall i \in \{2, 4, 8\}\\
\textrm{s.t.} \quad
    & \sum_{i} i \times N ^ {TP_i} \leq N \\
    & \sum_{i} (N^{TP_i} \times L^{TP_i}) \geq L \\
    & Performance^{TP_i, f_i}(L^{TP_i}) \leq SLO \\
    \end{aligned}
\label{eq1}
\end{equation}

This approach guarantees the energy optimal configuration.
However, it introduces non-negligible overheads (\emph{i.e.}, $\sim$100s of ms) due to the large search-space for the solver.
Hence, it cannot be used to select the correct system configuration at fine-grain intervals (\emph{e.g.}{}, every few seconds).
Next we show how to break the task into a hierarchy of subtasks and use an approximation heuristic to reduce the computation complexity.

\subsection{Hierarchical Control for Dynamic Load}
\label{sec:hierarchy}

\system{}
simplifies computations by assigning specific optimization tasks to individual controllers.
Instead of searching for a globally optimal configuration, controllers set locally optimal values for individual knobs under the constraints imposed by the upper-level controllers and under the assumption that the lower-level controllers operate 
at the highest performance configuration.
This 
allows the controllers to 
operate at varying time scales--from minutes for node adjustments down to seconds for frequency tuning.
The different scales are needed as each operational change involves distinct overheads and energy-efficiency impacts. 

\vspace{1pt}\noindent\textbf{Scale-out/in} 
At every epoch (\emph{e.g.}{}, 30 minutes), the cluster manager
computes the minimal number of nodes per pool that can support the load of a given request type.
The manager
uses a \emph{load predictor} to forecast the incoming load, $PL$, for each request type based on its historic data (\emph{e.g.}{}, via lightweight load templates~\cite{smartoclock}).
Moreover, the manager assumes that all instances
will run at the highest-performance configuration (\emph{i.e.}, TP8 at 1980 MHz).
Then,
if the predicted peak of a given request type is $PL$, and the maximum load that a node can support for this request type is $ML$, the manager assigns
{$\ceil[\big]{\frac{PL}{ML}}$} nodes to that pool.
By consolidating the work into a small number of nodes, the system tries to minimize the cost for the user and the idle energy of lightly-loaded GPUs.

\emph{Handling fragmentation:}
Allocating resources  to handle the peak loads can cause resource underutilization.
If overprovisioning accumulates across pools, the energy efficiency drops.
To prevent such cases, \system{} assigns one instance less to a given instance pool and moves the leftover load to the instance pool of the next larger request type.
The cluster manager uses this information to forward a fraction of the load for a given request type to the appropriate larger instance pool during the next scheduling epoch (\emph{e.g.}{}, 30 minutes).
In this way, only the instance pool for the largest requests can be overprovisioned minimizing the cluster-level fragmentation.

\vspace{1pt}\noindent\textbf{Shard-up/down}
At every epoch (\emph{e.g.}{}, every 5 minutes), the pool manager decides how 
to split the assigned $N$ GPUs from the cluster manager into correct model parallelism (how many instances to create in the pool) and tensor parallelism (how many GPUs to use for each instance).
The pool manager uses a simplified version of \Cref{eq1} assuming that all instances run at the highest GPU frequency (\emph{i.e.}, 1980 MHz).
Thus, the goal is to minimize the energy, while operating with the fixed number of GPUs running at the highest frequency, and controlling only the parallelism knob.

\emph{Accounting for the overheads:}
\system{} stores the transitioning overheads (scale-out/in, shard-up/down) in an \emph{Overhead Table}.
This table is integrated with the controllers,
so that when they calculate the energy benefits of new configurations, they can take into account the costs of reconfiguration.
The controllers evaluate whether the energy savings gained from re-configuring justify the associated overheads and downtime.

\emph{Reducing downtime:}
The reconfiguration cannot occur simultaneously on all instances due to the risk of long downtime.
Instead, \system{} employs a staggered reconfiguration approach, where a subset of instances is reconfigured (\emph{e.g.}{}, re-sharded) at a time. 
This ensures that while some instances are undergoing reconfiguration, others remain operational to handle ongoing workloads.
The system 
first reconfigures the instances that
have higher
potential energy savings.

\vspace{1pt}\noindent\textbf{Scale-up/down}
Finally, at every epoch (\emph{e.g.}{}, 5s), the instance manager fine tunes the GPU frequency for further energy savings given the assigned model parallelism.
The instance manager uses the performance profile
to first filter-out frequencies that violate the SLO under the current load.
Then, it uses the energy profile
to pick an acceptable frequency that minimizes the energy consumption.

\subsection{Reduced Overheads for Smooth Reconfiguration}
\label{sec:minoverheads}

To enable frequent reconfiguration, \system{} 
proposes a set of techniques to minimize the overheads of (1) \emph{scaling-in/out} the number of LLM inference servers, (2) \emph{sharding-up/down} the parallelism of a given instance, and (3) \emph{scaling-up/down} the GPU frequency of a given instance.

\vspace{1pt}\noindent\textbf{Scaling in/out inference servers}
\system{} reduces the overheads of creating a new server instance by implementing several strategies.
First,
it keeps the model weights cached locally within the cluster (shown in \Cref{fig:architecture}) avoiding the need to fetch them from a global repository.
Second, 
it
starts VMs from a snapshot with the entire state of the inference engine already initialized,  reducing the boot-up time.
This snapshot includes pre-loaded libraries, GPU drivers and inference engine configurations.
Third, it proactively creates new VMs in the background, outside of the critical path of active workload handling. 
Specifically, \system{} predicts the peak load for the next scheduling epoch and starts the extra VMs before the epoch starts.
By having these VMs ready to go, \system{} can seamlessly offload a fraction of the load to new instances without any noticeable latency impact.

\begin{figure}[t]
    \centering
    \includegraphics[width=\columnwidth]{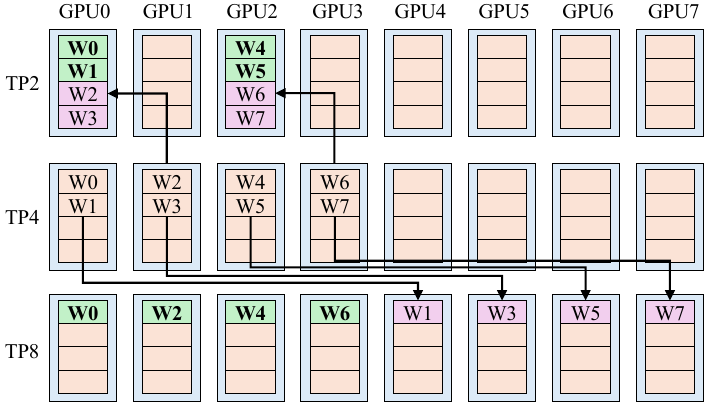}
    \caption{Example of re-sharding a TP4 model to TP2/TP8.}
    \label{fig:reshard_example}
\vspace{-4mm}
\end{figure}

\vspace{1pt}\noindent\textbf{Sharding up/down an instance}
To reduce the re-sharding overheads, \system{} optimizes the distribution of weights across GPUs.
We propose two techniques to minimize the data transfers and latency of individual transfers.
First, the system develops a graph matching algorithm that maximizes the amount of weights that remain stationary in their current GPUs.
The algorithm takes current weight distribution and desired tensor parallelism as inputs, and outputs the source and destination GPUs and fraction of weights to be transferred between each source-destination pair.
Specifically, the algorithm constructs a bipartite graph where nodes represent GPUs in the current and next configurations.
Edges between nodes represent potential transfers, weighted by the amount of data to be transferred.
Then, it applies a maximum weight matching algorithm to find the optimal transfer plan that minimizes the total weight of the edges (\emph{i.e.}, minimizes the amount of data transferred).
Second, to reduce the transfer latency, the system uses inter-GPU direct transfers via NVLink, 
allowing them to send fractions of their weights in parallel to other GPUs
without any host intervention.

\Cref{fig:reshard_example} shows an example of re-sharding from TP4 to TP2 and TP8.
Consider first the case when  going from a lower to a higher-level parallelism (TP4$\rightarrow$TP8).
In the initial state (TP4), GPUs 0-3 hold a quarter of the model weights each.
In the final state (TP8), all GPUs need to hold an eight of the model weight.
Thus, the first four GPUs already have their required state, and they only need to send half of their weights to the remaining four GPUs.
The four transfers (\emph{e.g.}, GPU0$\rightarrow$GPU4, GPU1$\rightarrow$GPU5,\ldots) happen in parallel.
Therefore, the re-sharding  requires the time to send 1/8 of the model weights via NVLink (around 50ms in our setup).

Consider now the case when going from a higher to a lower parallelism (TP4$\rightarrow$TP2).
In the final state (TP2), two GPUs need to hold a half of the weights each.
As each GPU initially holds a quarter of the weights, we merge the weights from two GPUs to a single one: GPU1 sends weights W2/W3 to GPU0, and GPU3 sends weights W6/W7 to GPU2.
As these two transfers happen in parallel, the total re-sharding time is the time to send 1/4 of the model weights (around 100ms in our setup).
\autoref{tab:reshard} shows the re-sharding overheads with different source/destination pairs with our optimized approach.
Some configurations quickly switch to other configurations (transition time 0), other changes incur larger overheads (transition time 4T, where T is the time to send 1/8 of the weights).

Moreover, on some transitions, the instance continues serving the requests with the same throughput.
This is the case when scaling from a smaller to a larger tensor parallelism (\emph{e.g.}{}, TP4$\rightarrow$TP8).
The old instance sends a fraction of the weights to the new instance, without increasing its memory footprint.
During other transitions, the current instance needs to operate under lower throughput.
This is the case when the instance scales from a larger to a smaller tensor parallelism (\emph{e.g.}{}, TP8$\rightarrow$TP4).
Some GPUs used by the old instance accept extra weights, reducing their memory capacity to serve new requests.
In general, whenever the GPU memory required to hold model weights increases, the throughput decreases due to the lower memory capacity for the incoming requests.

Finally, after the weights are sent to the correct memories, the inference engine needs to synchronize the GPUs that run the new instance.
State-of-the-art engines, such as vLLM~\cite{vllm}, perform this operation in a few 100s of milliseconds to a few seconds.
During this period the instance cannot receive any load, causing noticeable downtime.
To reduce the downtime, 
\system{} allows the old instance to process the requests while the new instance is going through the synchronization process. 
Only when the new instance comes online, the old instance is removed.
This is possible only when the sum of the memory from the old and new instances is below the GPU's memory capacity.
When the sum exceeds the memory capacity (\emph{e.g.}{}, TP2 and TP4 with 70B parameters), the old instance needs to be shutdown before the new instance is created.
Overall, different transitions incur different overheads and instance downtime requiring a fraction of the load to be shifted to another instance.
\system{} minimizes the overheads, and considers their impact on the overall efficiency when deciding whether to re-configure an instance.

\begin{table}[t]
\vspace{6pt}
\centering
\footnotesize
\begin{tabular}{ccccccc}
\toprule
Src/Dst & TP2 & 4TP2 & TP4 & TP2+TP4 & 2TP4 & TP8     \\
\midrule
TP2 & \cellcolor{green!30}0 & \cellcolor{red!50}4T & \cellcolor{orange!50}2T & \cellcolor{orange!50}2T & \cellcolor{orange!50}2T & \cellcolor{yellow!50}T\\
4TP2 & \cellcolor{green!30}0 & \cellcolor{green!30}0 & \cellcolor{green!30}0 & \cellcolor{green!30}0 & \cellcolor{green!30}0 & \cellcolor{green!30}0 \\
TP4 & \cellcolor{orange!50}2T & \cellcolor{orange!50}2T & \cellcolor{green!30}0 & \cellcolor{orange!50}2T & \cellcolor{orange!50}2T & \cellcolor{yellow!50}T \\
TP2+TP4 &  \cellcolor{green!30}0 & \cellcolor{orange!50}2T & \cellcolor{green!30}0 & \cellcolor{green!30}0 & \cellcolor{yellow!50}T & \cellcolor{yellow!50}T\\
2TP4 &  \cellcolor{yellow!50}T & \cellcolor{yellow!50}T & \cellcolor{green!30}0 & \cellcolor{yellow!50}T & \cellcolor{green!30}0 & \cellcolor{green!30}0 \\
TP8 & \cellcolor{yellow!50}T & \cellcolor{yellow!50}T & \cellcolor{yellow!50}T & \cellcolor{yellow!50}T & \cellcolor{yellow!50}T & \cellcolor{green!30}0\\
\bottomrule
\end{tabular}
\caption{Overhead of transferring model weights on a re-sharding.
T is the time unit to move 1/8 of the model (\emph{e.g.}{}, with 300GB/s NVLink bandwidth on NVIDIA DGX H100~\cite{h100} and Llama2-70B model~\cite{llama2}, $T=50ms$).
}
\label{tab:reshard}
\vspace{-3mm}
\end{table}

\vspace{1pt}\noindent\textbf{Scaling up/down the frequency}
The overheads of changing the GPU frequency are minimized by keeping the NVIDIA System Management Interface (\texttt{nvidia-smi}) monitor program directly loaded into memory. 
This eliminates the need to reload the program every time a frequency adjustment is required.
Moreover, by running the controller in privileged mode, \system{}
avoids the overhead associated with OS-user interactions, allowing for rapid frequency adjustments.

\subsection{Predictive Scheduling for Request Heterogeneity}
\label{sec:pools}

To map the heterogeneity of requests to the heterogeneous instance pools,
the cluster controller in \system{} uses an \emph{output-length predictor} to anticipate the request type and steer requests to the correct instance pool. 
The predictor acts as a proxy model that takes input prompt and classifies the output as short, medium or long.
Based on the predicted output length and known input length, the cluster manager forwards the request to the pool manager being in charge for a given request type.
If the instance pool is currently overloaded, the cluster manager forwards the request to the next available pool for a larger request type.
Once the request arrives to the correct pool, the pool manager needs to pick an instance from the pool.
Specifically, the manager uses the generated models from the profiling step to 
predict energy and response times 
of each instance after potentially adding a new request to that instance.
Then, it chooses the instance that minimizes total energy while staying within per-instance throughput determined by the SLO.

\vspace{1pt}\noindent\textbf{Handling mis-predictions}
If the system
over-estimates a request length, the request gets routed to a higher-performance pool.
Hence, it runs with sub-optimal energy, but its latency remains unaffected. Conversely, if a request length is under-estimated, the request is placed to a lower-performance pool, potentially missing its SLOs.
Similarly,
load mis-predictions can result in insufficient resources for a given pool during request bursts.
In both cases, due to some mis-predictions, the system needs to react to the created \emph{emergency event}.

When an instance manager detects that its queue is building up, indicating that the rate of request processing is lower than the rate of request arrival, it triggers an emergency event.
First, 
the instance manager tries to re-order the requests in its queue and prioritizes those requests that are about to miss their deadline.
Second, if some requests will miss their deadlines even after the reordering, the instance manager ramps up the frequency of its GPUs to the maximum value.
Third, if the backlog persists or worsens, the instance manager re-steers some requests that have not started their execution.
A subset of requests is moved to another instance within the pool via the pool manager. 
Finally, if all the attempts are insufficient, the instance manager resorts to more drastic measures. One such measure involves squashing requests that have been waiting for processing beyond a specified threshold.
This action signals users to retry their requests, allowing the frontend system to redirect these requests to alternative instance pools or retry them later when system load has stabilized.
\subsection{\system{} Implementation}
\label{sec:impl}

We build \system{} on top of vLLM~\cite{vllm}, a state-of-the-art LLM inference platform.
However, \system{}'s modularity allows it to be integrated with other
platforms, \emph{e.g.}, TensorRT-LLM~\cite{tensorRTLLM}, without modifications.
We implement controllers as lightweight gRPC servers  with low memory and compute requirements.
Cluster and pool managers are hosted  in a dedicated VM for robust management.
Instance managers are co-located with the LLM inference engine instances for low communication overheads.
For output length prediction, we leverage a BERT-based proxy model~\cite{aiops2024qiu}, which provides accurate and efficient classification of incoming requests.
For load prediction, we use a template-based approach that uses historical data to model load patterns over a week~\cite{smartoclock}. 
The pool manager employs Python's PuLP library~\cite{pulp} for solving MILP.
\system{} models energy and performance using the \emph{interp1d} function from the SciPy~\cite{scipy} Python library.

\begin{figure}[t]
    \centering
    \includegraphics[width=\columnwidth]{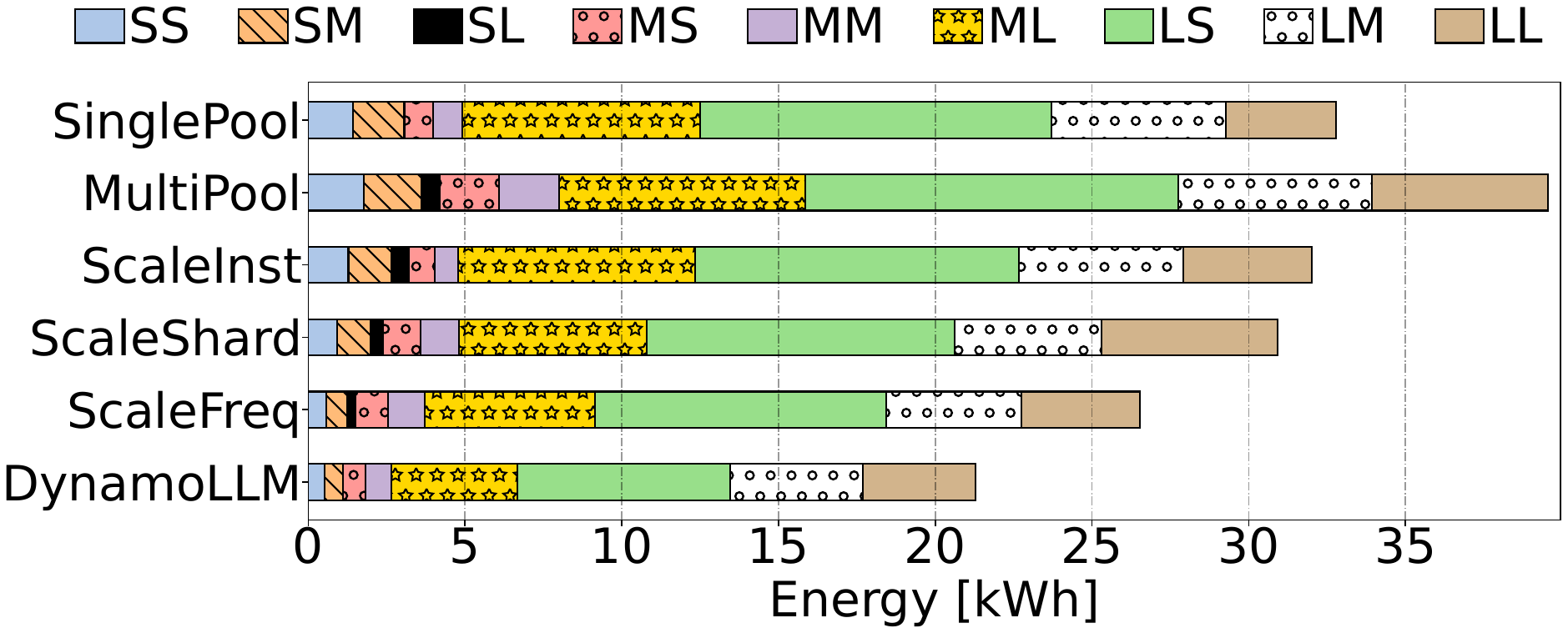}
    \caption{Energy consumption with the six evaluated systems running open-source Llama2-70B model~\cite{llama2} with 1-hour open-source production traces~\cite{splitwise}.}
    \label{fig:eval_energy_cluster}
    \vspace{-4mm}
\end{figure}

\section{Evaluation}

\subsection{Evaluation Setup}
\label{sec:methodology}

We run our experiments on servers with 8 H100 GPUs~\cite{h100}.
We show the results for Llama2-70B~\cite{llama2}, but other models (\emph{i.e.}, Mixtral~\cite{mixtral2}, Falcon~\cite{falcon}, BLOOM~\cite{scao2022bloom}) follow the same trends.
We set the load using production-level traces:
1 hour open-source traces~\cite{splitwise} and 1-day and 1-week traces for \emph{Coding} and \emph{Conversation} from our fleet.
We compare \system{} to five systems.
\emph{SinglePool} (a state-of-the-practice baseline) schedules all the requests to the common pool of instances running with TP8 at the highest GPU frequency.
\emph{MultiPool} separates LLM instances in multiple per-request-type pools.
\emph{ScaleInst}, \emph{ScaleShard}, and \emph{ScaleFreq} additionally scale the number of instances in the pool, model parallelism, or GPU frequency according to the current load, respectively.

\subsection{Cluster-Level Experiments}
\label{sec:cluster}

We first evaluate the system on a cluster of GPU servers using the 1h open source production traces for the \emph{Conversation} service~\cite{splitwise}.
We provision the baselines with 12 H100 servers to handle the peak load,
while \system{} scales the number of servers according to the current load.

\vspace{1pt}\noindent\textbf{Energy}
\Cref{fig:eval_energy_cluster} shows the energy consumption of the cluster for the experiment.
MultiPool increases the energy consumption by 20\% over SinglePool, because it allocates a larger number of resources while always operating at the highest-performance configuration. Meanwhile, ScaleInst, ScaleShard, ScaleFreq and \system{} reduce the energy consumption by 4.1\%, 7\%, 19\% and 35\%, respectively.
ScaleInst/Shard/Freq reduce the energy by configuring one knob but leave substantial space for further savings.
Finally, \system{} synchronously scales multiple knobs to achieve the lowest energy consumption.
We further breakdown the total energy per request type.
\Cref{fig:eval_energy_cluster} shows that longer requests (\emph{e.g.}, LL) and highly-popular requests (\emph{e.g.}, ML) consume dis-proportionally more energy than the other types.

\begin{figure}[t]
    \vspace{0mm}
    \centering
    \includegraphics[width=\columnwidth]{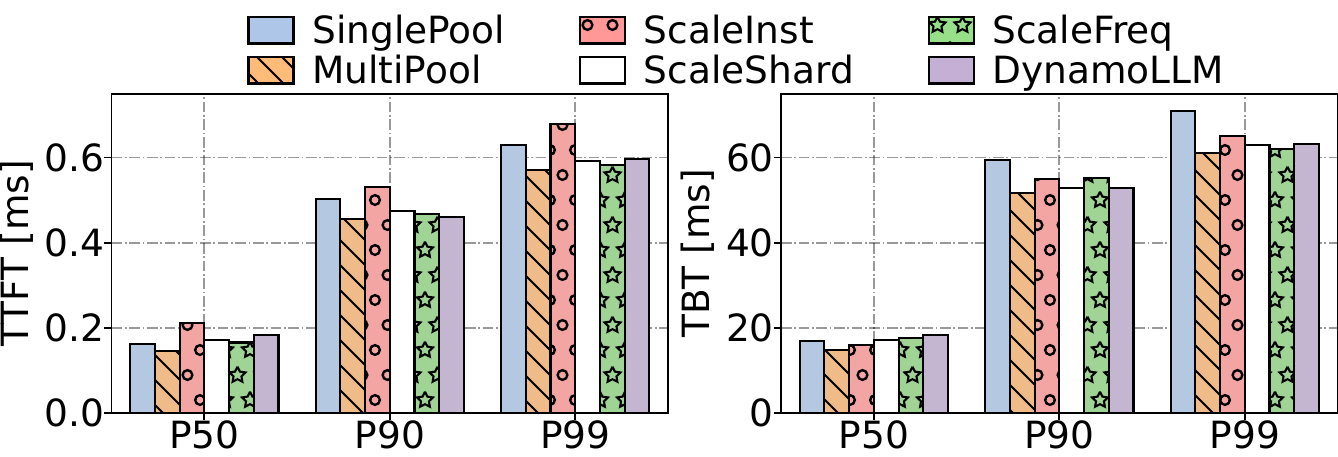}
    \caption{Summary of the latencies for each of the systems running open-source Llama2-70B model~\cite{llama2} with 1-hour open-source production traces~\cite{splitwise}.
    }
    \label{fig:eval_perf_cluster}
    \vspace{-2mm}
\end{figure}

\begin{figure}[t]
    \centering
    \includegraphics[width=\columnwidth]{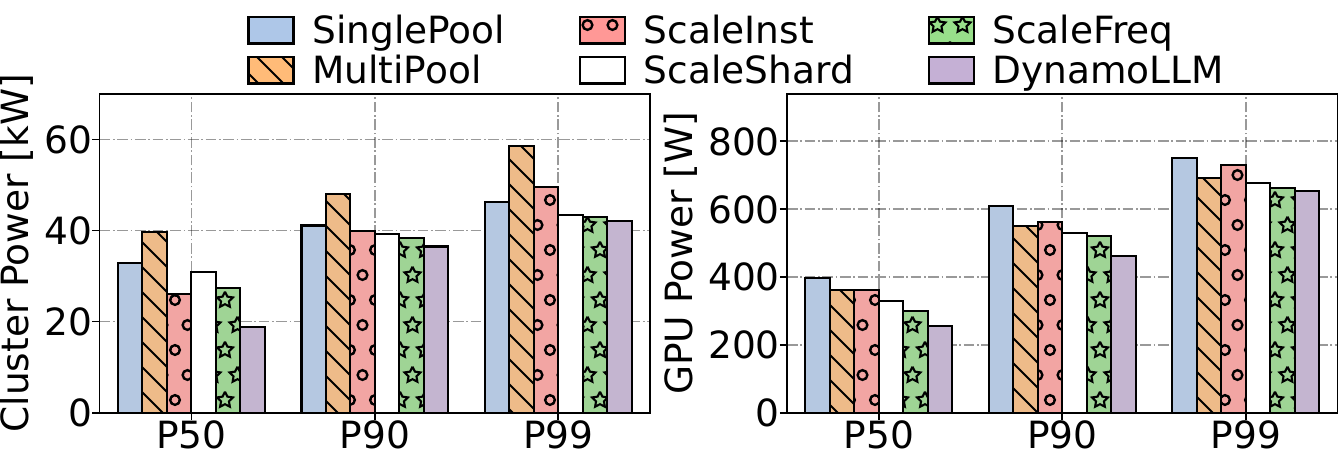}
    \caption{Summary of the power for each of the systems running open-source Llama2-70B model~\cite{llama2} with 1-hour open-source production traces~\cite{splitwise}.
    }
    \label{fig:eval_power_cluster}
    \vspace{-4mm}
\end{figure}

\vspace{1pt}\noindent\textbf{Latency}
\Cref{fig:eval_perf_cluster} shows the 
TTFT/TBT latencies for each system.
By separating request types into different resource pools, MultiPool 
removes the head-of-line blocking effect and reduces the latencies over SinglePool.
Similarly, ScaleShard and ScaleFreq, and \system{} reduce the tail latency.
However, these systems slightly increase the P50 latency by operating in lower-performance modes when there is available SLO slack.
On the other hand, ScaleInst increases the tail latency due to the large overheads of creating a new inference server on the critical path of users' load.
Overall, \system{} reduces the P99 TTFT and TBT latencies by 5.3\% and 11.1\% over SinglePool, respectively,
while it increases the P50 TTFT and TBT latencies by
11.4\% and 7.6\%, respectively.

\vspace{1pt}\noindent\textbf{Power}
\Cref{fig:eval_power_cluster} shows the
power consumption across the systems for the cluster (right figure) and on average per-GPU (left figure).
Due to operating in energy-efficient modes, \system{} effectively reduces both cluster and per-GPU power. 
\system{} reduces the P50 and P99 power consumption over the baseline by 43\% and 9\%, respectively.

\begin{figure}[t]
    \centering
    \includegraphics[width=\columnwidth]{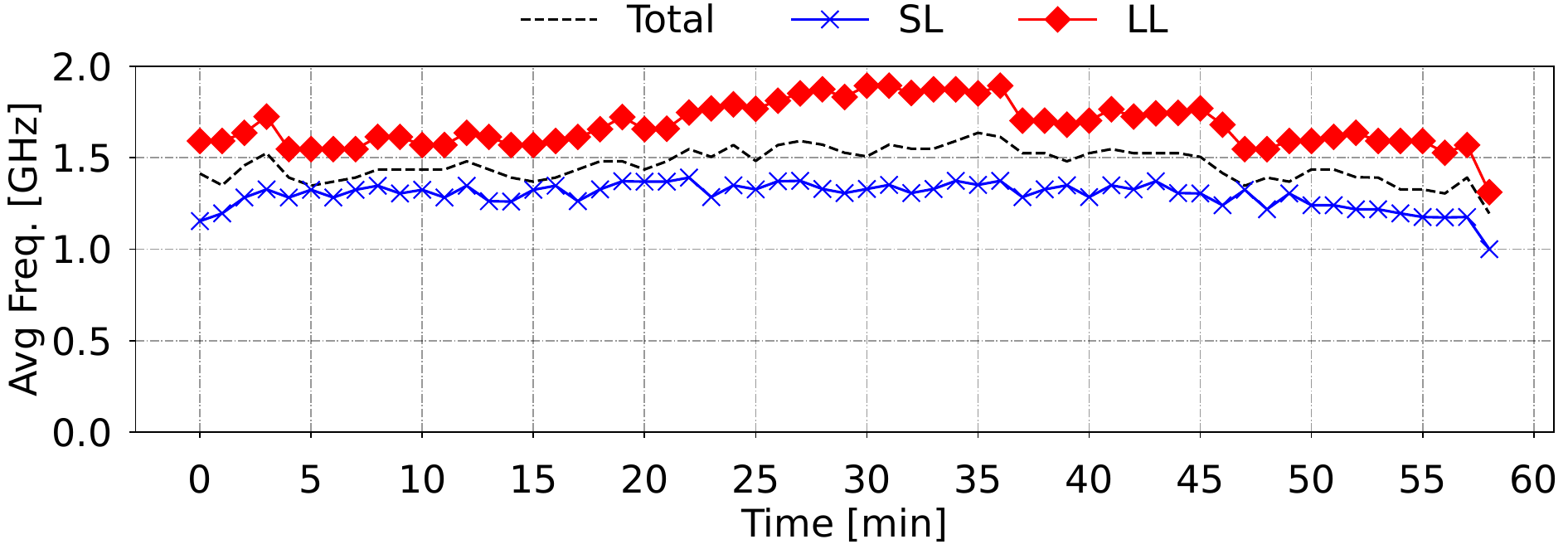}
    \caption{GPU Frequency over one hour for \system{} running open-source Llama2-70B model~\cite{llama2} with 1-hour open-source production traces~\cite{splitwise}.
    }
    \label{fig:eval_freq_cluster}
    \vspace{-1mm}
\end{figure}

\begin{figure}[t]
\centering
\includegraphics[width=\columnwidth]{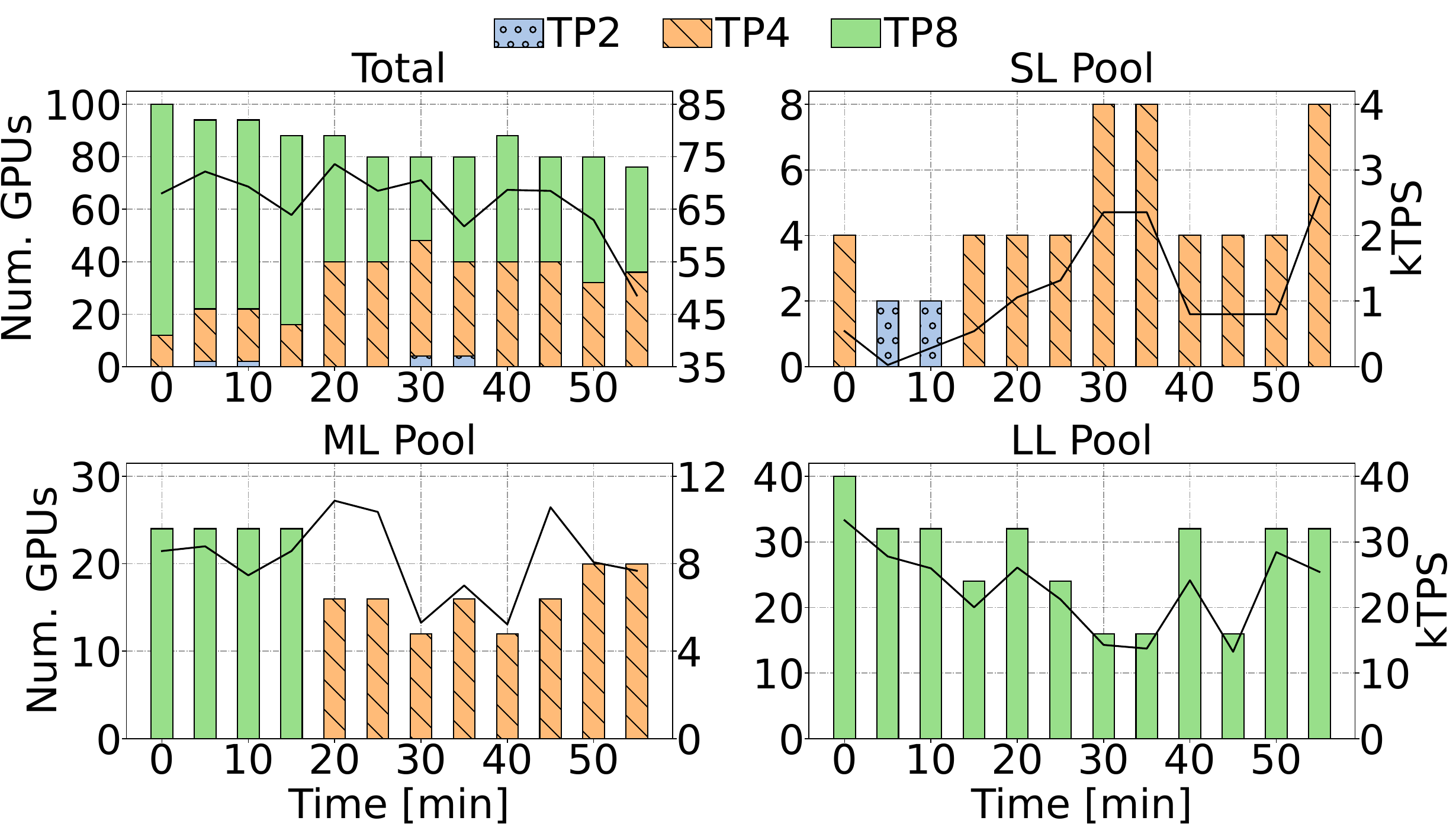}
\caption{Number of GPUs used for each sharding configuration (TP2, TP4 or TP8) over time running open-source Llama2-70B model~\cite{llama2} with 1-hour open-source production traces~\cite{splitwise}.}
\label{fig:eval_shard_cluster}
\vspace{-3mm}
\end{figure}

\vspace{1pt}\noindent\textbf{Frequency changes}
\Cref{fig:eval_freq_cluster} shows the average GPU frequency over time for
(1) the whole cluster,
(2) the pool serving short requests, and
(3) the pool serving long requests.
Average frequency is always significantly lower than the maximum allowed frequency (1980 MHz) that is used by the baseline.
\system{} effectively accommodates different request types by operating their pools at different frequencies.

\vspace{1pt}\noindent\textbf{Sharding changes}
\Cref{fig:eval_shard_cluster} shows the number of GPUs used for different model parallelisms (TP2, TP4 and TP8) for the whole cluster and for the individual pools (SL, ML and LL).
The figure also shows the load over time that a given pool experiences.
Different pools operate with different model parallelisms and \system{} efficiently changes the parallelism as the load changes.

\subsection{Sensitivity Studies}

\begin{figure}[t]
\centering
\includegraphics[width=\columnwidth]{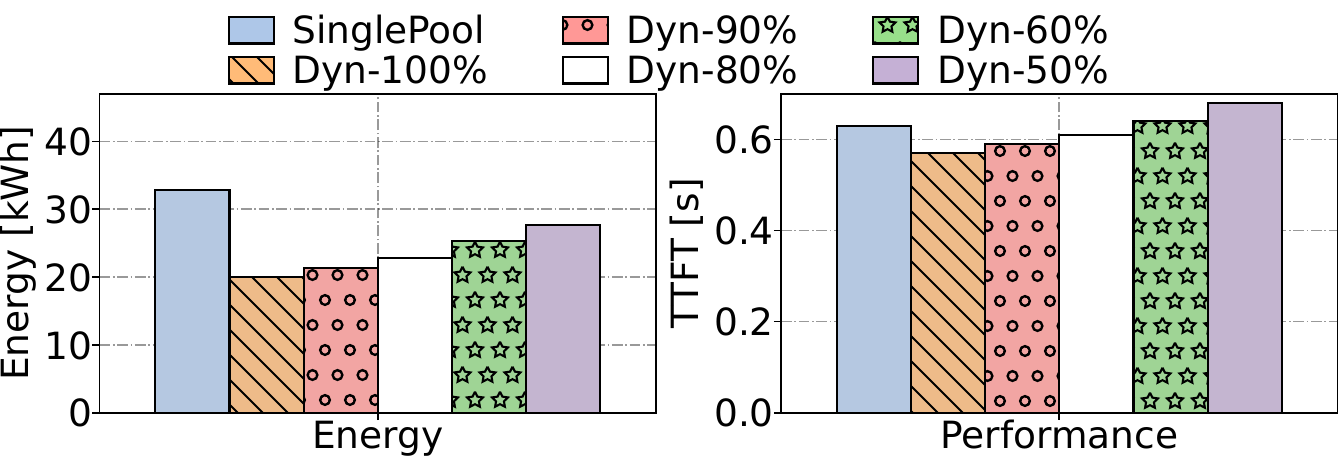}
\caption{Energy and performance with different accuracy running open-source Llama2-70B model~\cite{llama2} with 1-hour open-source production traces~\cite{splitwise}.}
\label{fig:eval_sens_accuracy}
\vspace{-2mm}
\end{figure}

\vspace{1pt}\noindent\textbf{Sensitivity to predictor accuracy}
We analyze how the accuracy of the prediction models affects the overall system efficiency.
We introduce bounded errors for the output length misclassification and measure the energy consumption with medium load.
\Cref{fig:eval_sens_accuracy} shows that the impact of the predictor accuracy is modest for both energy and performance.
Compared to an environment with no error, an environment with an 40\% error increases the energy consumption by 13\% and TTFT by 7.3\%.
The reason for robustness to prediction errors is that \system{} can promptly detect mis-predictions and re-configures the knobs accordingly.

\begin{figure}[t]
    \centering
    \includegraphics[width=\columnwidth]{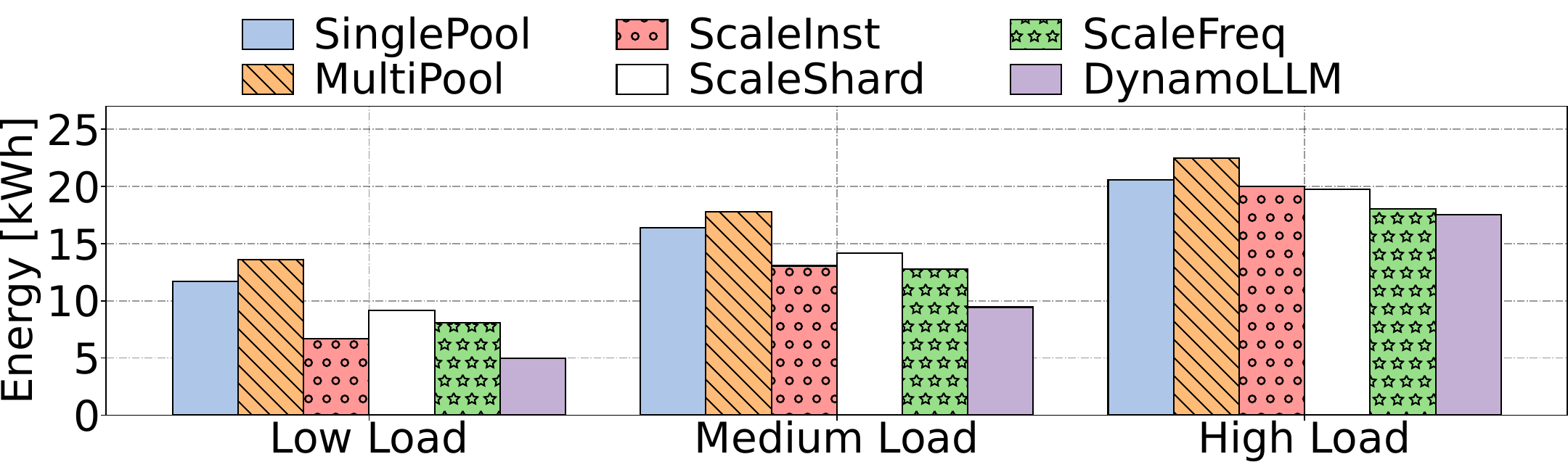}
    \caption{Energy with different levels of load running open-source Llama2-70B model~\cite{llama2}.}
    \label{fig:eval_sens_load}
    \vspace{-2mm}
\end{figure}

\vspace{1pt}\noindent\textbf{Sensitivity to load}
We evaluate \system{} with different system loads.
We generate Low, Medium, and High loads with a Poisson distribution for request inter-arrival times.
\Cref{fig:eval_sens_load} shows the energy consumption of the five evaluated systems with different load levels.
With Low, Medium, and High load, \system{} reduces the energy of SinglePool baseline by 51\%, 40\%, and 23.4\%, respectively.
As the load increases, the energy savings of \system{} reduce, because the system more frequently needs to operate at higher frequencies with higher levels of model parallelism.

\begin{figure*}
    \begin{minipage}{\columnwidth}
\centering
\includegraphics[width=\columnwidth]{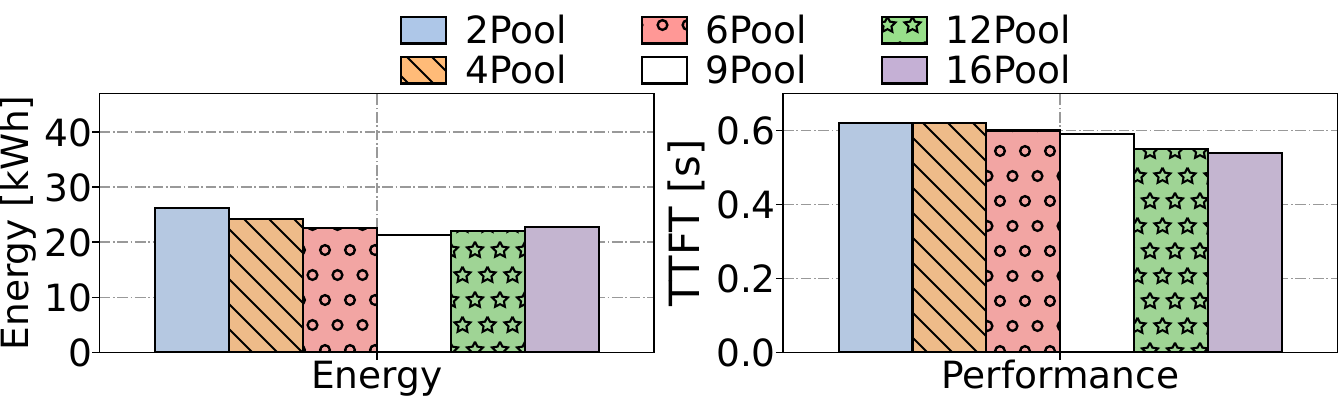}
\caption{Energy and performance with different number of pools (or request types) running open-source Llama2-70B model~\cite{llama2} with 1-hour open-source production traces~\cite{splitwise}.}
\label{fig:eval_sens_poolcnt}
\end{minipage}
\hspace{3mm}
\begin{minipage}{\columnwidth}
    \centering
    \includegraphics[width=\columnwidth]{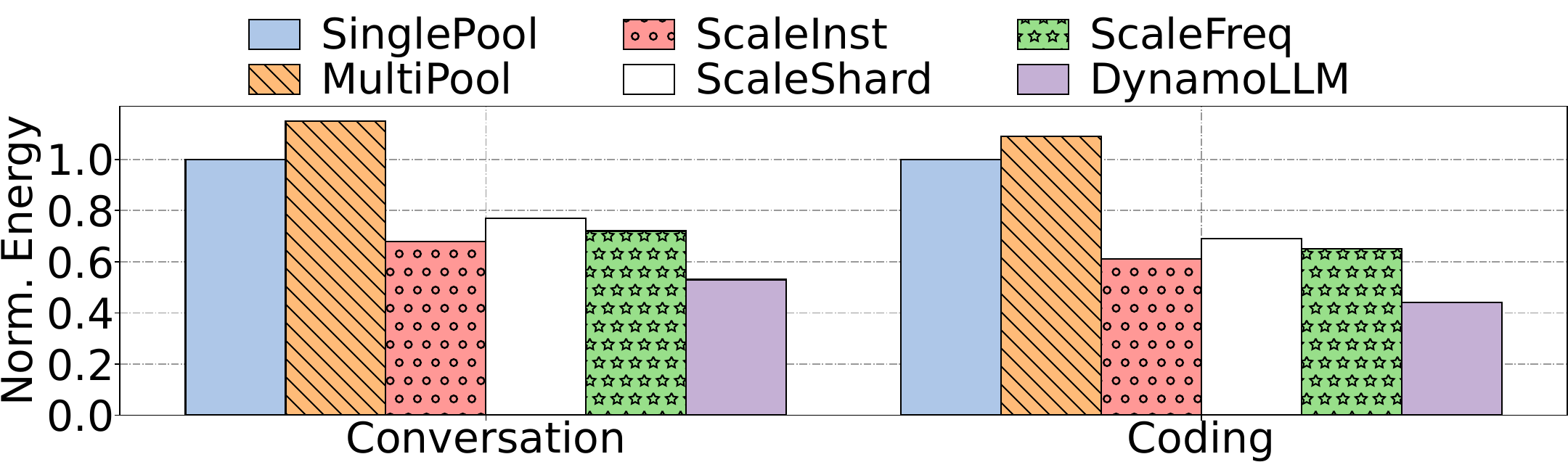}
    \caption{Normalized energy consumption for the six evaluated systems with the week-long production traces running open-source Llama2-70B model~\cite{llama2}.}
    \label{fig:eval_sim_energy}
\end{minipage}
\end{figure*}

\vspace{1pt}\noindent\textbf{Sensitivity to number of pools}
\Cref{fig:eval_sens_poolcnt} shows the energy consumption and performance (TTFT) of \system{} with different number of request pools.
Recall that our chosen design has 9 pools.
By adding too many pools (12 or 16), the system gets fragmented, and the idle energy of GPUs results in the overall energy increase.
Reducing the number of pools (2 or 4) prevents the system from fine tuning the frequency and the model parallelism for specific request types. The performance improves by adding moderately more pools because it helps remove head of the line blocking and introduces more resources for execution. 

\begin{figure*}
\begin{minipage}{\columnwidth}
    \centering
    \includegraphics[width=\columnwidth]{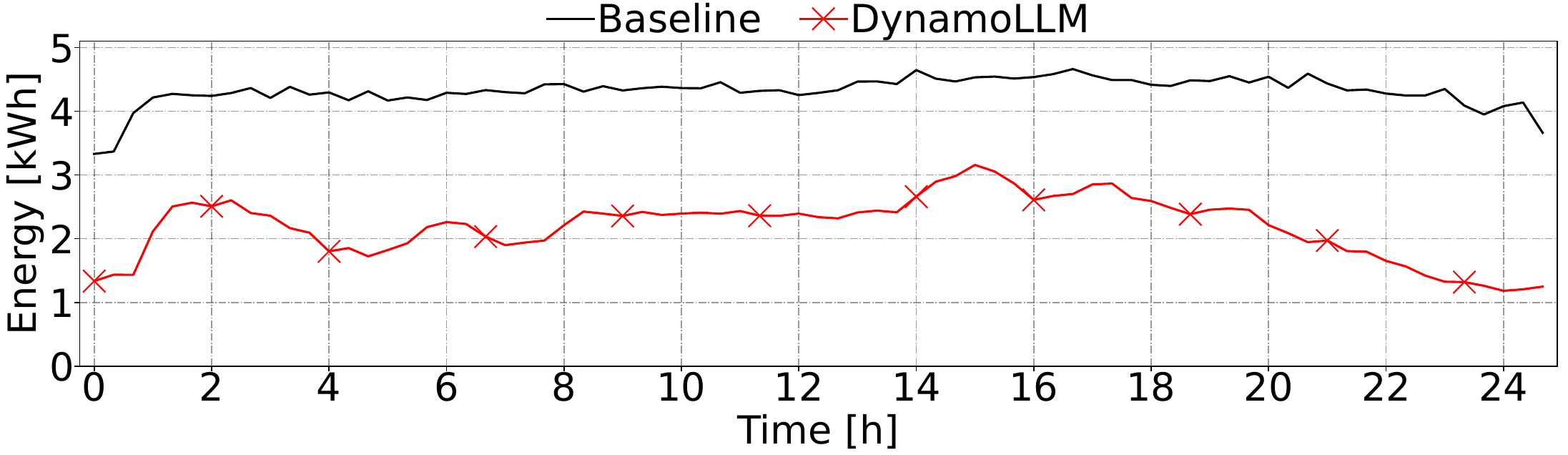}
    \caption{Energy consumption of SinglePool and \system{} with 1-day production traces running open-source Llama2-70B model~\cite{llama2}.
    }
    \label{fig:long}
\end{minipage}
\hspace{3mm}
\begin{minipage}{\columnwidth}
    \centering
    \includegraphics[width=\columnwidth]{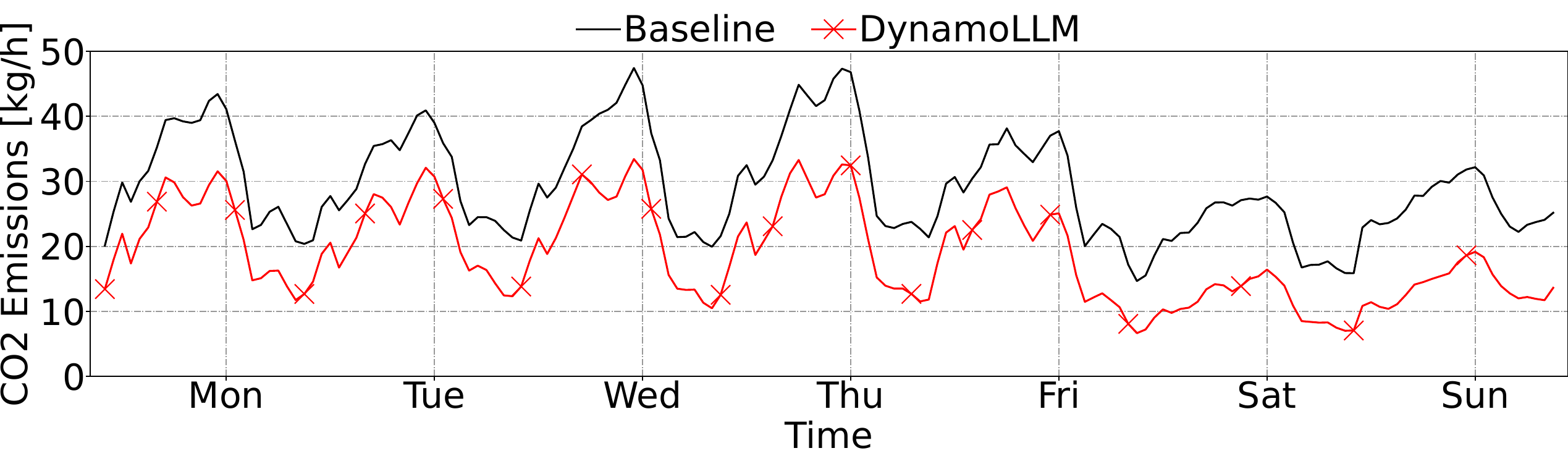}
    \caption{Carbon emissions over time for the week-long conversation traces with \system{} and SinglePool baseline running open-source Llama2-70B model~\cite{llama2}.}
    \label{fig:eval_sim_co2}
\end{minipage}
\vspace{-3mm}
\end{figure*}

\subsection{Long Cluster-Level Experiments}
\label{sec:emul}

We run longer experiments by running the 1-day traces for the \emph{Conversation} service.
The trace covers all invocations for a subset of the service's instances during a typical work day.
We run the experiment for 24-hours on 11 H100 servers with SinglePool and scale the number of instances based on the load in \system{}.
\Cref{fig:long} shows the energy consumption over 5-minute intervals for the two systems.
\system{} reduces the energy consumption over the baseline during peak hours (when dynamic power dominates), and during the low utilization times (when idle power dominates).
Over the whole day, \system{} reduces the energy consumption by 42\%.

\subsection{Large-Scale Simulations}
\label{sec:sim}

To generalize our insights to large-scale, we develop a discrete-time simulator that simulates the energy consumption of different systems using production traces.
\Cref{fig:eval_sim_energy} shows the normalized energy consumption for the five evaluated systems using 1-week traces for \emph{Conversation} and \emph{Coding} services.
\system{} significantly reduces the energy consumption for both types of services.
\system{} operates in higher energy-efficient modes for the \emph{Conversation} service due to its typically shorter input lengths (ML dominant request type).
On the other hand, the \emph{Coding} service has deep valleys during the night and weekends.
Thus, \system{} exploits the periods of low load to save energy.
\system{} reduces the energy consumption over the baseline by 47\% and 56\% for the \emph{Conversation} and \emph{Coding} services,
respectively.

\subsection{Cost and Carbon Emission}

\vspace{1pt}\noindent\textbf{User cost}
\system{} reduces the operational cost for users
by minimizing the number of GPUs and optimizing their energy efficiency.
The number of GPU servers for the week-long experiments reduces from 40 to 24.6 on average (38.5\% cost reduction).
Using the current GPU VM pricing~\cite{gpuCost},
this saves \$1362.7/hour.
By reducing the energy consumption, \system{} reduces the associated energy costs by up to 56\%.
As energy costs~\cite{energyCost} are currently substantially lower than GPU costs, this translates to only \$4.4/hour savings.

\vspace{1pt}\noindent\textbf{Carbon emissions}
The energy consumption translates into the amount of operational $CO_2$ emissions.
We use the traces of carbon intensity~\cite{watttime} for a week-long period from multiple grids and map the carbon intensity to the energy consumption over time for the SinglePool baseline and \system{}.
\Cref{fig:eval_sim_co2} shows the operational carbon emissions over time for the two systems for CAISO~\cite{caiso}.
SinglePool and \system{} consume 5t and 3.1t/week of $CO_2$.
These substantial savings (38\%) make a step towards sustainable LLM environments.

\section{Related Work}
\label{sec:related}

\vspace{1pt}\noindent\textbf{Cluster resource and power management}
A rich body of work seeks to improve resource efficiency under the SLO constraints through resource management for a wide range of latency sensitive workloads, such as microservices~\cite{sinan} and DL workloads, through effective resource sharing~\cite{parties, clite, twig}, dynamic allocation~\cite{antman}, and hardware reconfiguration~\cite{cuttlesys}. Others
focus on approaches that enable safe power management and oversubscription~\cite{polca, thunderbolt, smoothOperator} leveraging workload characteristics~\cite{predictionVM, flexDatacenter} and system state~\cite{smartoclock}. 

\vspace{1pt}\noindent\textbf{Energy-efficient workloads}
Prior works focused on energy-efficiency for CPU workloads~\cite{ecofaas, twig, eetl, pegasus}, and researchers started exploring unique energy properties of GPU workloads~\cite{e_energy, cloudEnergy, wordstowatts}. 
Recent schemes build on top and 
manage energy and power consumption for DNN inference and training~\cite{zeus, alert, dynamictraining} through frequency scaling~\cite{batchDVFS, geepfas, batchsizer, heterDVFS, edgebert, llmDVFS, indicator_directed}, autoscaling~\cite{autoscale}, and resource partitioning and mapping~\cite{wattwiser, edgebert}. We show that improving energy efficiency for LLM inference necessitates a comprehensive view of all available knobs. \system{} is holistic framework that dynamically reconfigures all the knobs considering the diversity and dynamism of requests and loads.

\vspace{1pt}\noindent\textbf{Efficient LLM inference serving}
Recent works propose approaches to improve LLM inference efficiency through heterogeneous resources~\cite{alizadeh2024llm} and platforms~\cite{spotserve}, memory and key-value cache management~\cite{vllm, flashattention}, and node- and cluster-level scheduling~\cite{pets, exegpt, splitwise, orca, sarathi, paella, alpaserve}. While these studies focus on improving throughput or latency, we show that optimizing energy efficiency for LLM inference exhibits distinct trade-offs between performance, energy consumption, and overheads and thus requires a comprehensive framework.

\section{Conclusion}
\label{sec:conclusion}

We present \system{}, the first energy-management framework for LLM inference clusters. 
\system{} exploits heterogeneity in inference compute properties and fluctuations in inference workloads to save energy. 
The system automatically and dynamically configures the energy-optimal organization of the cluster (number of instances, model parallelism and GPU frequency) while performing under performance guarantees.
\system{} reduces energy, carbon emissions and cost to the customer by 53\%, 38\% and 61\%, respectively.


\bibliographystyle{IEEEtranS}
\bibliography{refs}

\end{document}